\documentclass[letterpaper]{article} 
\usepackage{aaai25}  
\usepackage{times}  
\usepackage{helvet}  
\usepackage{courier}  
\usepackage[hyphens]{url}  
\usepackage{graphicx} 
\usepackage{natbib}  
\usepackage{caption} 
\usepackage{algorithm}
\usepackage{algorithmic}

\usepackage[T5]{fontenc}
\usepackage[utf8]{inputenc}

\usepackage{times}
\usepackage{latexsym}
\usepackage{amsmath}
\usepackage{tcolorbox}

\usepackage{hhline}
\usepackage{colortbl}
\usepackage{inconsolata}
\usepackage{ragged2e}
\usepackage{color}
\usepackage{array,multirow,graphicx,booktabs}
\usepackage{arydshln}

\usepackage{subfig}

\usepackage{amsmath,amssymb,amsfonts}
\usepackage{algorithmic}
\usepackage{graphicx}
\usepackage{textcomp}
\usepackage{rotating}
\usepackage{tabularray, tabularx}
\usepackage{xcolor,pifont}
\usepackage{tikz}
\usetikzlibrary{fadings}
\usepackage{adjustbox}
\usepackage{soul}
\usepackage{xurl}
\usepackage{longtable}
\usepackage{microtype}

\usepackage{newfloat}
\usepackage{listings}
\DeclareCaptionStyle{ruled}{labelfont=normalfont,labelsep=colon,strut=off} 
\floatstyle{ruled}
\newfloat{listing}{tb}{lst}{}
\floatname{listing}{Listing}
\newcommand*\colourcheck[1]{%
  \expandafter\newcommand\csname #1check\endcsname{\textcolor{#1}{\ding{52}}}%
}
\colourcheck{blue}

\newcommand*\xmarkcheck[1]{%
  \expandafter\newcommand\csname #1check\endcsname{\textcolor{#1}{\ding{55}}}%
}
\xmarkcheck{red}

\title{ViFactCheck: A New Benchmark Dataset and Methods for Multi-domain News Fact-Checking in Vietnamese}
\author{
    Thai-Hoa Tran\textsuperscript{\rm 1,2},
    Quang-Duy Tran\textsuperscript{\rm 1,2},
    Khanh Quoc Tran\textsuperscript{\rm 1,2},
    Kiet Van Nguyen\textsuperscript{\rm 1,2}\thanks{Corresponding author.}
}
\affiliations{
    \textsuperscript{\rm 1} Faculty of Information Science and Engineering, University of Information Technology\\

    \textsuperscript{\rm 2} Vietnam National University, Ho Chi Minh City, Vietnam\\
    \{21522082,20512013\}@gm.uit.edu.vn, \{khanhtq,kietnv\}@uit.edu.vn
}

\begin{document}

\maketitle

\begin{abstract}
The rapid spread of information in the digital age highlights the critical need for effective fact-checking tools, particularly for languages with limited resources, such as Vietnamese. In response to this challenge, we introduce ViFactCheck, the first publicly available benchmark dataset designed specifically for Vietnamese fact-checking across multiple online news domains. This dataset contains 7,232 human-annotated pairs of claim-evidence combinations sourced from reputable Vietnamese online news, covering 12 diverse topics. It has been subjected to a meticulous annotation process to ensure high quality and reliability, achieving a Fleiss Kappa inter-annotator agreement score of 0.83. Our evaluation leverages state-of-the-art pre-trained and large language models, employing fine-tuning and prompting techniques to assess performance. Notably, the Gemma model demonstrated superior effectiveness, with an impressive macro F1 score of 89.90\%, thereby establishing a new standard for fact-checking benchmarks. This result highlights the robust capabilities of Gemma in accurately identifying and verifying facts in Vietnamese. To further promote advances in fact-checking technology and improve the reliability of digital media, we have made the ViFactCheck dataset, model checkpoints, fact-checking pipelines, and source code freely available on GitHub. This initiative aims to inspire further research and enhance the accuracy of information in low-resource languages\footnote{\url{https://github.com/QuangDiy/ViFactCheck}}.
\end{abstract}

\section{Introduction}
The rapid proliferation of digital information has created significant challenges in distinguishing between accurate and false information. The spread of disinformation, rumors, and fake news has become a global concern with far-reaching consequences for individuals, societies, and public discourse. As noted by \citet{lazer2018science}, the extensive spread of fake news can have severe negative impacts on individuals and society. It can cause confusion and misunderstanding, disrupt social order, and even threaten national security. 

\begin{figure}[ht]
\centering
\begin{tikzpicture}
    
    \draw[gray, rounded corners=6pt] (0,0) rectangle (7.5,3);
   
    \draw[gray, rounded corners=6pt] (0,-0.1) rectangle (7.5,-7.2);

    \node at (3.75, 1.6) {
        \begin{minipage}{7cm}
        \small
        \includegraphics[height=15pt,width=15pt]{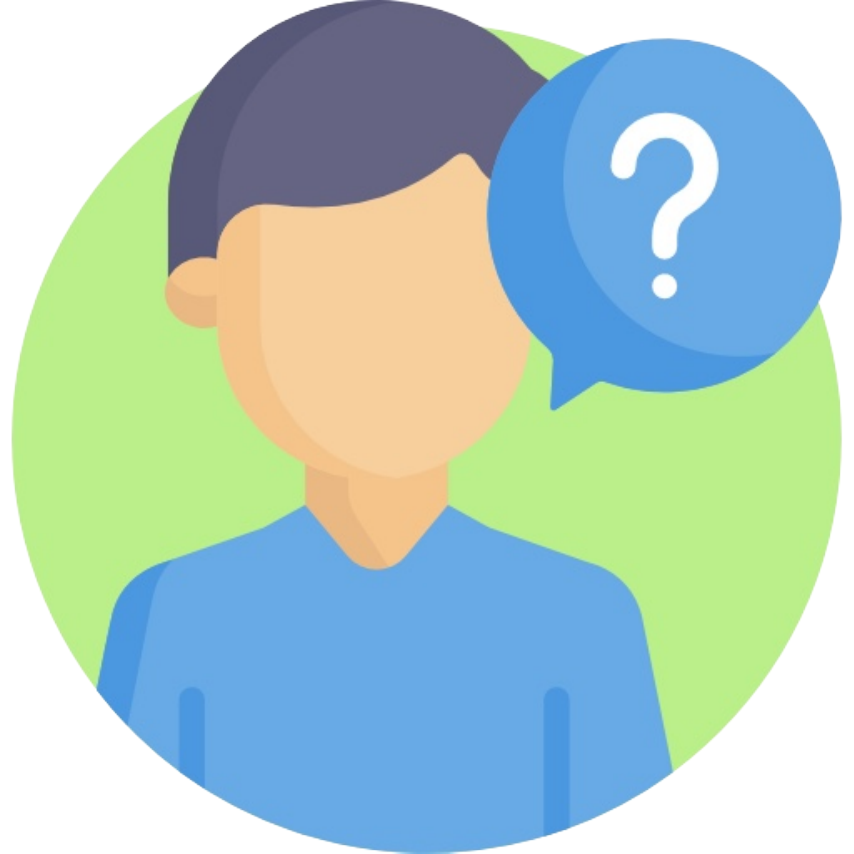}
        \textcolor{teal}{\textbf{Claim:}}\\
        Các công dân trẻ tiêu biểu cũng tham gia vào giải chạy bộ ``Bước chân xanh'' nhằm hưởng ứng chiến dịch Giờ Trái đất năm 2023. \\ 
       \textbf{English:} Exemplary young citizens also participate in the ``Green Steps'' running event to support the Earth Hour campaign in 2023.
        \end{minipage}
    };
    \node at (3.75, -3.65) {
        \begin{minipage}{7cm}
        \small
        \includegraphics[height=15pt,width=15pt]{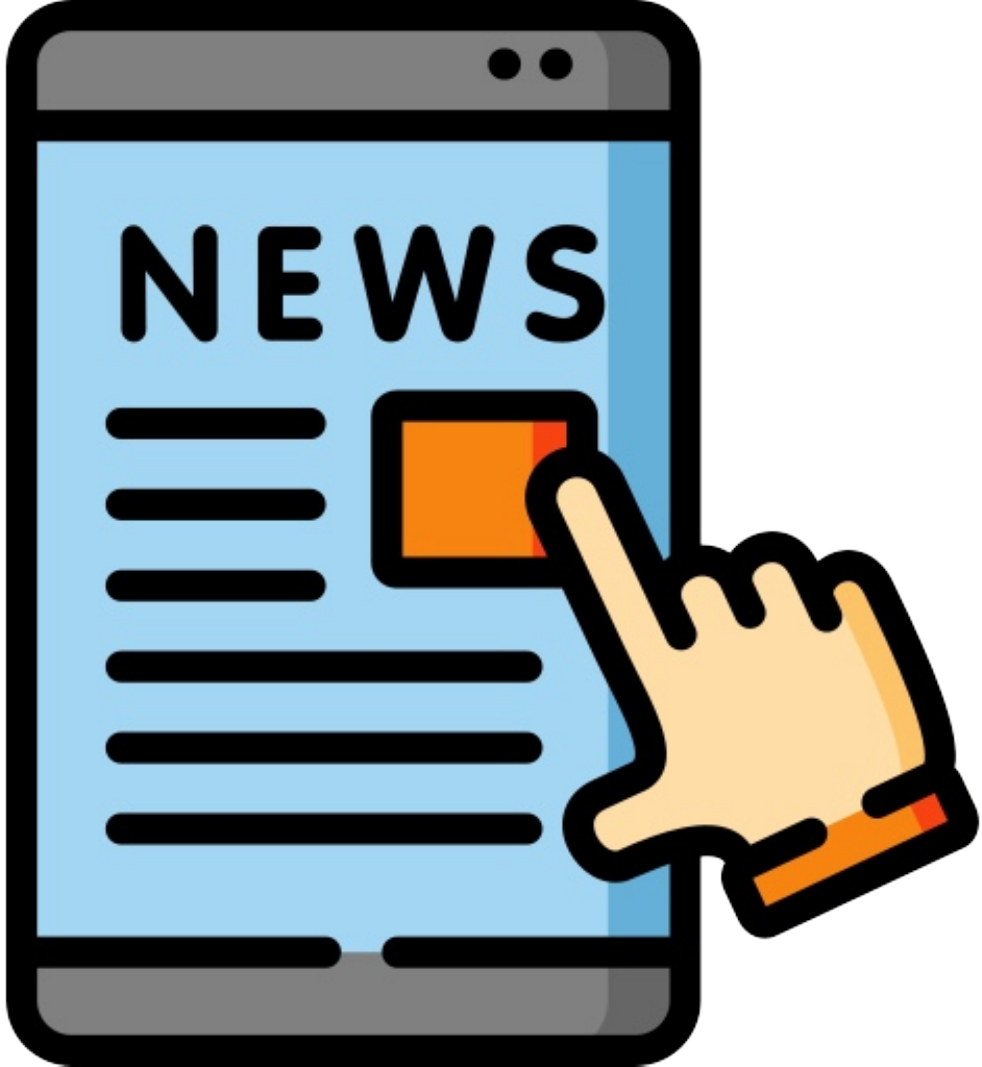}
        \textcolor{teal}{\textbf{Context:}} \\
    TPO-Sáng 25/3, Thành Đoàn, Hội LHTN Việt Nam TPHCM , Hội Sinh viên Việt Nam TPHCM tổ chức Giải chạy bộ “Bước chân xanh” lần thứ 2. Giải chạy thu hút hơn 1.000 người tham gia \textcolor{cyan}{hưởng ứng chiến dịch Giờ Trái đất năm 2023}. Bên cạnh đông đảo đoàn viên, thanh niên, sinh viên, \textcolor{cyan}{giải chạy bộ “Bước chân xanh” còn thu hút các gương công dân trẻ tiêu biểu TPHCM}, các hoa hậu, á hậu, văn nghệ sĩ trẻ... cùng tham gia.\\
    \textbf{English:} TPO-March 25th, the HCM Youth Union and the Vietnam National Union of Students in HCM City organized the 2nd ``Green Steps'' running event. The race attracted over 1,000 participants in \textcolor{cyan}{response to the Earth Hour campaign in 2023}. In addition to a large number of union members, youth, and students, \textcolor{cyan}{the ``Green Steps'' running event also attracted notable young citizens of HCM City}, beauty queens, runners-up, young artists, and others to participate.

        \end{minipage}
    };
    
    \draw[gray, rounded corners=5pt, fill=white] (2.75,0.2) rectangle (4.75,-0.3);

    \node[align=center] at (3.75,-0.05) {\textcolor{blue}{\textbf{Support}} \bluecheck }; 
\end{tikzpicture} 
\caption{An example of the Vietnamese fact-checking task. Words \textcolor{cyan}{highlighted in blue} represent key evidence used to support the classification of the claim as ``Supported''.}
\label{figure:example}
\end{figure} 

Fact-checking, a rigorous process to verify the accuracy of claims in specific contexts, relies on informed individuals using evidence, reasoning, and available information to make well-founded judgements. Figure \ref{figure:example} provides a specific illustration for Vietnamese fact-checking. Although substantial efforts have been devoted to fact-checking in English \cite{thorne-etal-2018-fever,aly2021feverous,schuster2021vitamin}, resources for fact-checking in low-resource languages like Vietnamese are limited. This scarcity primarily stems from the limited availability of guidance resources to analyze the structure and semantics of Vietnamese.

To bridge this gap, this study presents the development of ViFactCheck, the first publicly available human-curated fact-checking benchmark tailored to multiple domains Vietnamese news. Our main contributions are described as follows:

\begin{enumerate}
    \item \textbf{Dataset Construction:} We developed ViFactCheck, a comprehensive dataset encompassing 12 critical domains of Vietnamese online news. This dataset contains 7,232 rigorously vetted human-annotated claims, thereby ensuring a robust foundation for both research and practical applications.
    \item \textbf{Model Experimentation:} We utilized fine-tuning and prompting techniques to several state-of-the-art language models using the ViFactCheck dataset to assess their effectiveness in verifying information within the Vietnamese context. Our study includes fine-tuning and zero-shot in-context learning on both pre-trained and large language models, specifically adapted to this linguistic framework, to evaluate their efficacy.
    \item \textbf{In-depth Analysis}: Through detailed examinations of the challenges faced during the creation of the dataset and subsequent experimentation, this study offers profound insights into the hurdles of developing fact-checking systems for low-resource languages, guiding future advancements in the field.
\end{enumerate}

The remainder of this paper is structured as follows. Section \ref{fundamental} delves into the fundamentals of fact-checking tasks. Section \ref{dataset} describes the process of constructing the ViFactCheck benchmark dataset. Section \ref{experimental} discusses the results of our experiments and identifies key challenges encountered. Section \ref{conclusion} concludes with a summary of our findings and suggests directions for future research.

\section{Fundamental of Fact-Checking}\label{fundamental}

\begin{table*}[ht]
\centering
\begin{adjustbox}{width=.92\textwidth}
\begin{tblr}{
  column{4} = {r},
  column{3} = {c},
  column{5} = {c},
  cell{2}{1} = {r=5}{},
  cell{7}{1} = {r=5}{},
  hline{1} = {-}{0.08em},
  hline{2} = {-}{0.08em},
  hline{7,12} = {-}{},
}
 & \textbf{Dataset} & \textbf{Labels} & \textbf{\# Claims} & \textbf{Annotated Evidence} & \textbf{Language} & \textbf{Source} &\textbf{\#RS}\\ 
\begin{sideways}English\end{sideways} & FEVER \shortcite{thorne-etal-2018-fever} & 3 & 185,445 & \bluecheck & English & Wikipedia & Multi \\
& FEVEROUS \shortcite{aly2021feverous} & 3 & 87,026 & \bluecheck & English &  Wikipedia & Multi \\
 & VitaminC \shortcite{schuster2021vitamin} & 3 & 488,904 & \redcheck & English &  Wikipedia & Single\\
 & MultiFC \shortcite{augenstein-etal-2019-multifc}  & 2-40 & 36,534 & \bluecheck & English &  Fact-check & Multi\\
  & LIAR \shortcite{wang2017liar} & 6 & 12,836 & \redcheck  & English & Fact-check & W/O\\
\begin{sideways}Non-English\end{sideways} & CHEF \shortcite{hu2022chef} & 3 & 10,000 & \bluecheck & Chinese & News/Fact-check & Multi \\
 & DANFEVER \shortcite{norregaard-derczynski-2021-danfever} & 3 & 6,407 & \bluecheck & Danish & Wikipedia & Multi \\
 & ANT \shortcite{khouja-2020-stance} & 2 & 4,547 & \redcheck & Arabic & News & Multi \\
& ViWikiFC \shortcite{le2024viwikifc} & 3 & 20,976 & \redcheck & Vietnamese & Wikipedia & Single \\

& \textbf{ViFactCheck (Ours)} & 3 & 7,232  & \bluecheck & Vietnamese & News~ & Multi

\end{tblr}
\end{adjustbox} 
\caption{Comparative overview of typical open-domain fact-checking datasets. The type of Reasoning Steps (\#RS) column reflects the complexity involved in verifying the claims in each dataset.}
\label{table:comparative_benchmarks}
\end{table*}

\subsection{Foundational Benchmark Datasets}
Benchmark datasets are crucial in the development and evaluation of fact-checking algorithms, serving as the foundation upon which these systems are tested and fine-tuned. The FEVER \citep{thorne-etal-2018-fever} is particularly notable, containing more than 185,000 claims sourced from Wikipedia, each meticulously annotated with evidence to support or refute the claims. Following FEVER, the FEVEROUS dataset \citep{aly2021feverous} extends these capabilities by incorporating not only text but also structured data such as tables and lists, presenting a more comprehensive dataset that challenges algorithms to parse and verify information across different formats. Another significant dataset, MultiFC \cite{augenstein-etal-2019-multifc}, compiles claims from 26 different fact-checking websites, covering various topics and offering a rich environment to test the adaptability of verification systems to different contexts and types of misinformation. These benchmark datasets play a critical role in advancing the field of fact-checking, providing a diverse set of challenges and inspiring the development of diverse open-domain fact-checking datasets in many languages \cite{schuster2021vitamin,wang2017liar,hu2022chef,norregaard-derczynski-2021-danfever,khouja-2020-stance}. The comparison of multi-domain fact-checking datasets is summarized in Table \ref{table:comparative_benchmarks}.

\subsection{Advanced Methods in Fact-Checking}
The evolution of fact-checking methods has significantly advanced through the adoption of sophisticated machine learning technologies. Notably, the use of Pre-trained Language Models (PLMs) and Large Language Models (LLMs) like BERT and other transformer-based architectures \cite{devlin2019bert} has been instrumental. These models, leveraging deep learning, are highly effective in processing and analyzing the context within texts based on patterns in data, making them exceptionally effective for tasks such as evidence retrieval and claim verification \cite{nie2019combining,soleimani2019bert,liu-etal-2020-fine}. By fine-tuning these models on specific fact-checking datasets, researchers can adapt their capabilities to better recognize and interpret the nuances of misinformation. Furthermore, researchers have explored prompting techniques with these models to direct their focus without extensive retraining, enhancing their utility in diverse applications \cite{huang-etal-2023-zero,pan-etal-2023-fact}. The synergy of language models with traditional retrieval and verification methods has also given rise to hybrid models, which combine the depth and adaptability of machine learning with the precision of rule-based systems \cite{vlachos-riedel-2014-fact}, graph modeling \cite{zhong-etal-2020-reasoning}, leading to more robust and accurate fact-checking solutions.

\subsection{Vietnamese Research on Fact-Checking}
Research within Vietnam on fact-checking has been making significant strides, particularly with the development of customized datasets that address the unique linguistic characteristics of Vietnamese \cite{duong2023fact,le2024viwikifc}. A notable study by \citet{duong2023fact} has produced a dataset with more than 129K triples checked for fact, specifically designed to evaluate the effectiveness of fact-checking algorithms under Vietnamese linguistic constraints. This approach not only enhances the precision of fact-checking in Vietnam but also contributes significantly to the global body of knowledge. It showcases how fact-checking technologies can be adapted to different linguistic and cultural contexts, providing a model for similar adaptations in other regions.

\section{Dataset Creation Process}\label{dataset}

\begin{figure}[ht]
  \centering
  \includegraphics[width=0.45\textwidth]{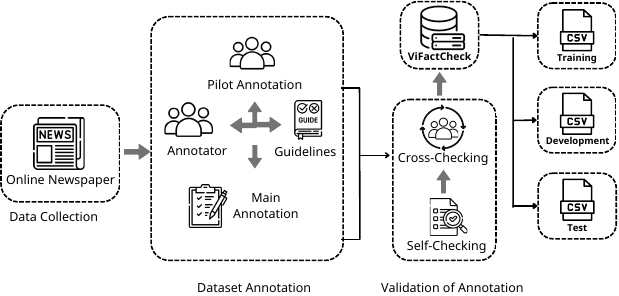}
 \caption{The ViFactCheck dataset contruction process.}
 \label{figure:3_dataset}
\end{figure}

Figure \ref{figure:3_dataset} shows the development of ViFactCheck, the first multi-domain Vietnamese news fact-checking benchmark. The dataset construction included three phases: data collection, dataset annotation, and annotation validation, each rigorously monitored by experts to ensure dataset quality.

\subsection{Data Collection}\label{Data Collection}
This research constructs a dataset from articles sourced from nine licensed and widely-read Vietnamese online newspapers, detailed in the Appendix B. These sources were chosen for their comprehensive and timely news coverage, ensuring the relevance and reliability of the dataset. We extracted datasets that included titles, content, topics, lead descriptions, and URLs of articles published between February and March 2023. The selection of this period aims to capture the current dynamics of news reporting, providing a contemporary snapshot of media trends.

The initial corpus contained 1,000 articles covering 12 topics. Notably, news leads were merged with their respective contents to form a ``Full Context'' field, thereby enriching the dataset with a more comprehensive narrative view. This methodological rigor ensure the utility of dataset in advancing research on media analysis and computational linguistics.

\subsection{Dataset Annotation}\label{Dataset Annotation}
\begin{figure}[ht]
    \centering
    {\includegraphics[width=0.21\textwidth]{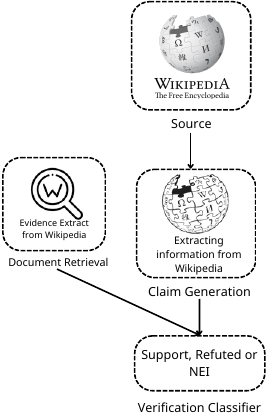}} 
    {\includegraphics[width=0.21\textwidth]{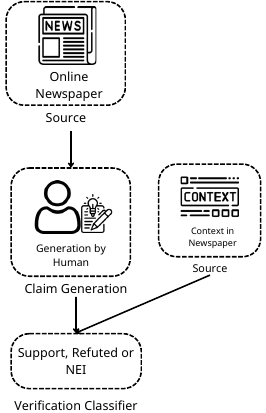}} 
    \subfloat[\label{fig:fever}  \citet{thorne-etal-2018-fever}.]{\hspace{.5\linewidth}}
    \subfloat[\label{fig:Vifact} Our proposed process.]{\hspace{.5\linewidth}}
    \caption{Comparison of the labeling pipelines in the FEVER and ViFactCheck datasets.}
\end{figure}

The construction methodology proposed for Vietnamese news differs from the conventional methods in previous datasets \cite{khouja-2020-stance,norregaard-derczynski-2021-danfever}, which mimic the FEVER approach (Figure \ref{fig:fever}). Recognizing the nuanced and dynamic nature of online news, our method employs human annotators to extract and interpret contextual nuances and factual details from news articles (Figure \ref{fig:Vifact}). This human-centered approach enhances the naturalness and relevance of the data, enabling the dataset to better represent complex real-world information scenarios.

By assigning labels that reflect the context of each article, our methodology supports intricate inference tasks that require analysis across multiple pieces of evidence. This refined approach ensures that our dataset is exceptionally well-suited for advanced fact-verification systems, significantly contributing to the accuracy and effectiveness of misinformation detection in the digital media landscape.

\subsubsection{Pilot Annotation} is used to familiarize the annotators with the claim generation and verification process described above. Seven native Vietnamese-speaking university students were involved as annotators. We conducted a pilot annotation with each annotator annotating 120 claims corresponding to 20 random articles. Annotators were instructed to proofread each claim carefully and rigorously in accordance with the annotation guidelines. Details of the annotators recruitment and guidelines can be found in Appendices C and D.

To verify the integrity of the pilot annotation process, we conducted thorough reviews of both the claims and their corresponding labels. The expert provided detailed feedback and asked the annotators to review any details or labels that did not meet the requirements of the annotation guidelines. 

\subsubsection{Main Annotation} 
Following a pilot phase that familiarized the annotators with the tasks, each was assigned a specific subset to ensure focused and deep engagement. Throughout this phase, strict adherence to established guidelines was paramount to ensure consistency and enhance the overall quality of the dataset. 

\textbf{Claim Generation:} Before generating any claims, annotators conducted a thorough review of the article. This meticulous process ensures a deep understanding of the multiple facets of the article, facilitating an accurate interpretation of the information. Annotators then employed their expertise to construct claims that align with the predefined labels: Support, Refute, and NEI (Not Enough Information). Such rigorous adherence to these guidelines is essential for generating contextually relevant claims, thereby enhancing the reliability of the dataset and its utility in advancing fact-checking.

\textbf{Evidence Annotation:} In terms of evidence annotation, the task extends beyond simple identification. Annotators are required to meticulously annotate the supporting evidence for each claim derived from the phrases previously collected from the articles. To enhance the complexity of the dataset and the challenge it presents, annotators are instructed not to limit their claims to single pieces of evidence. Instead, they are required to craft intricate claims that amalgamate multiple pieces of evidence (Appendix F). This process involves breaking down the claim, collating diverse evidences, and performing multi-step reasoning. The ability to synthesize complex evidence not only enriches the data but also crucially underpins more sophisticated analyses.

\subsection{Validation of Annotation} \label{Validation of Annotation}

After completing the main annotation phases, we implemented several strategies to ensure the quality and consistency of the dataset: \textbf{(1)} Self-checking: Annotators review their own claims and labels, checking for grammatical errors and typographical mistakes. \textbf{(2)} Cross-checking: Annotators verify the work of their peers. Any identified errors are collaboratively discussed and corrected.

\textbf{Metric For Inter-Annotator Agreement}: Fleiss Kappa is widely used to evaluate inter-annotator agreement (IAA) in several tasks and is considered a benchmark for such measurements \cite{mchugh2012interrater, thorne-etal-2018-fever}. Consequently, we utilized the Fleiss Kappa metric \cite{fleiss1971measuring} to assess inter-annotator agreement, thus ensuring quality assurance in human annotation.

We randomly selected 10\% of the claims (n = 726) from the labeled dataset, assigning them to a group of three annotators. These claims, originally authored by different individuals, were relabeled without revealing the existing annotations. The inter-rater agreement was then calculated using the Fleiss Kappa measure. We achieved an agreement level of 0.83, indicative of a very high level of agreement among annotators, which confirms the high quality and reliability of our dataset.

\subsection{Words overlap and Semantic similarity analysis}
To evaluate the complexity of inference within our dataset, we employed two principal metrics: word overlap and semantic similarity. For word overlap, we used metrics including Longest Common Sequence (LCS), New Word Ratio (\%) (NWR), Jaccard Similarity (\%) (JS), and Lexical Overlap. For semantic similarity, we utilized the concept of Related Words, generating embeddings with SBERT \shortcite{reimers-2019-sentence-bert} and calculating correlations using cosine similarity. The results are summarized in Table \ref{table:LCS}. 

\begin{table}[ht]
\centering
\begin{adjustbox}{width=\linewidth}
\begin{tblr}{
  row{1} = {c},
  cell{2}{1} = {r=3}{},
  cell{5}{1} = {r=3}{},
  cell{2-7}{3-7} = {r},
  vline{3} = {-}{},
  hline{1,8} = {-}{0.08em},
  hline{2,5} = {-}{},
  hline{3-4,6-7} = {2-7}{dashed},
}
 &  & \textbf{LSC} & \textbf{NWR} & \textbf{JS} & \textbf{LO} & \textbf{RW}\\
Context & Support & 20.60 & 6.54 & 11.46 & 20.13 & 36.24\\
 & Refute & 18.10 & 11.50 & 10.06 & 17.90 & 34.00\\
 & NEI & 19.89 & 11.81 & 10.96 & 18.50 & 32.85\\
Evidence & Support & 17.70 & 17.13 & 63.52 & 73.87 & 86.89\\
 & Refute & 15.46 & 25.47 & 54.63 & 66.69 & 81.41\\
 & NEI & 16.71 & 26.84 & 57.56 & 64.39 & 81.13
\end{tblr}    
\end{adjustbox}
\caption{Relationship between claim-context and claim-evidence in the ViFactCheck dataset.}
\label{table:LCS}
\end{table}

\citet{mccoy-etal-2019-right} demonstrated that models face difficulties with low overlap ratios, necessitating advanced inference capabilities. Our dataset features claim-context pairs with minimal word overlap and semantic similarity, complicating model inference. In contrast, a strong correlation between claim-evidence pairs significantly enhances the performance of models when the appropriate evidence is retrieved. Further detailed analysis can be found in the Appendix G.

\section{Experiment and Results}\label{experimental}
\subsection{Baseline models}
Drawing on the transformative impact of transformer-based models in prior fact-checking studies \cite{thorne-etal-2018-fever,hu2022chef,norregaard-derczynski-2021-danfever}, our research employs pre-trained language models (PLMs) that utilize the BERT \cite{devlin2019bert} and RoBERTa \cite{liu2019roberta} architectures. Our study incorporates four specific models to address the fact-checking task, including two multilingual models mBERT \shortcite{devlin2019bert} and XLM-R \shortcite{conneau-etal-2020-unsupervised} and two monolingual models PhoBERT \shortcite{nguyen2020phobert} and ViBERT \shortcite{bui-etal-2020-improving} tailored to handle linguistic nuances effectively.

Moreover, the recent strides in large language models (LLMs) have solidified their utility in demonstrating robust contextual comprehension and reasoning capabilities, particularly in tasks that require deep understanding, such as fact-checking. Accordingly, our experimental framework includes four SOTA open-source LLMs designed for optimized performance in low-resource settings: Llama \shortcite{touvron2023llama}, Gemma \shortcite{team2024gemma}, and Mistral \shortcite{jiang2023mistral}. These models are pivotal in our methodology, providing a comprehensive approach to evaluating their effectiveness across diverse linguistic contexts.

\subsection{Software and Hardware Configurations} \label{config}
We employed the AdamW optimizer for fine-tuning pre-trained language models, as detailed by \citet{loshchilov2019decoupled}. The settings for these models included a learning rate of 5e-06, a dropout rate of 0.3, a batch size of 16, and a training duration of 10 epochs. Additionally, for PhoBERT, we segmented the text data using VnCoreNLP \cite{vu-etal-2018-vncorenlp}, adhering to the recommendations by \citet{nguyen2020phobert}. The dataset was partitioned into training, development, and test sets with a split ratio of 6:2:2, ensuring a balanced evaluation across all stages of model development.

For LLMs, we utilized the Unsloth framework with supervised fine-tuning using LoRA adaptation. The hyper-parameters were configured with a Lora rank of 16, Lora alpha of 16, a learning rate of 2e-04, a batch size of 16, and 5 epochs. All experiments were conducted on a RTX 4090 GPU with 24GB of memory, utilizing PyTorch version 2.2.1 and Transformers version 4.41.2, and took a total of five days to complete. Details of the models, prompt templates and parameters can be found in Appendices H, I and J.

\subsection{Main Results}\label{main_results}
Table \ref{tab:main_results} presents a detailed comparison of language models in fact-checking, examining their performance across different methods such as fine-tuning and prompting, and their efficiency in using Full Context versus Gold Evidence. Using the macro-average F1 score (\%), the analysis provides insights into the capabilities of the models, highlighting the strengths and limitations of each approach in processing complex information sets.

\begin{table}[ht]
\centering
\resizebox{.98\linewidth}{!}{%
\begin{tabular}{lccr}
\hline
\multicolumn{1}{c}{\textbf{Model}} & \textbf{Full Context} & \textbf{Gold Evidence} & \multicolumn{1}{c}{\textbf{$\Delta$}} \\ \hline
\multicolumn{4}{c}{\textbf{\textit{Fine-tuning PLMs}}} \\ \hline
PhoBERT$_{base}$ & 68.55 & 77.76 & \textcolor{blue}{$\uparrow$9.21} \\
PhoBERT$_{large}$ & 62.93 & 79.76 & \textcolor{blue}{$\uparrow$\underline{16.83}} \\
ViBERT & 59.95 & 72.18 & \textcolor{blue}{$\uparrow$12.23} \\
mBERT & 58.07 & 69.94 & \textcolor{blue}{$\uparrow$11.87} \\
XLM-R$_{base}$ & 65.40 & 81.10 & \textcolor{blue}{$\uparrow$15.70} \\
XLM-R$_{large}$ & \underline{75.42} & \underline{88.02} & \textcolor{blue}{$\uparrow$12.60} \\ \hline
\multicolumn{4}{c}{\textbf{\textit{Fine-tuning LLMs}}} \\ \hline
\textbf{Gemma} & \textbf{\underline{85.94}} & \textbf{\underline{89.90}} & \textcolor{blue}{$\uparrow$3.96} \\
Mistral & 70.13 & 88.63 & \textcolor{blue}{$\uparrow$18.50} \\
Llama2 & 41.47 & 79.53 & \textcolor{blue}{$\uparrow$\textbf{\underline{38.06}}} \\
Llama3 & 79.65 & 88.67 & \textcolor{blue}{$\uparrow$9.02} \\ \hline
\multicolumn{4}{c}{\textbf{\textit{Prompting LLMs}}} \\ \hline
Gemini & \underline{76.26} & \underline{74.88} & \textcolor{red}{$\downarrow$1.38} \\
Gemma & 45.05 & 39.47 & \textcolor{red}{$\downarrow$5.58} \\
Mistral & 61.02 & 57.31 & \textcolor{red}{$\downarrow$3.71} \\
Llama2 & 63.54 & 51.64 & \textcolor{red}{$\downarrow$\underline{11.90}} \\
Llama3 & 65.21 & 63.10 & \textcolor{red}{$\downarrow$2.11} \\ \hline
\end{tabular}%
}

\caption{Performance comparison of baseline models on the ViFactCheck test set. Context and Evidence indicate the use of Full Context and Gold Evidence, respectively, for Claim Verification. The best scores are highlighted in bold; models that outperform other peers are underlined. Performance differences ($\Delta$) are statistically significant ($p$ < 0.01), confirming robust gains or reductions when Full Context is employed compared to Gold Evidence.}
\label{tab:main_results}
\end{table}

\begin{figure*}[ht]
    \centering
    \includegraphics[width=.9\linewidth]{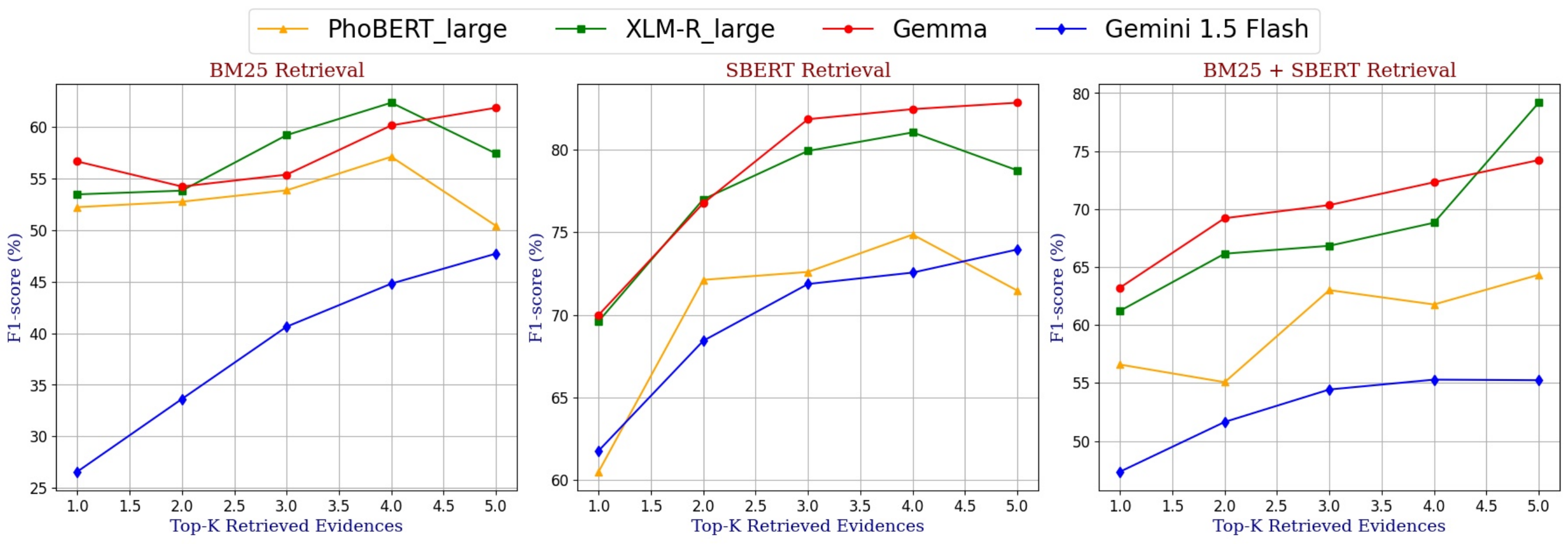}
    \caption{Comparative performance of various text retrieval models across different Top-K settings.}
    \label{fig:evidences_ir}
\end{figure*}

\textbf{Fine-tuning Pre-trained Language Models}
Among the PLMs, XLM-R$_{large}$ stands out with exemplary performance, scoring 75.42\% in Context and 88.02\% in Evidence. These results suggest that the scale and design of XLM-R$_{large}$ provide a robust model capable of handling the complexities inherent in determining the veracity of claims based on the provided contexts and evidence. Additionally, variants of BERT-based models also demonstrate considerable gains, with PhoBERT$_{large}$ in particular showing a significant leap in context understanding compared to its peers.

\textbf{Fine-tuning Large Language Models}
The LLMs, particularly Gemma, display remarkable effectiveness, outperforming other models in both Context (85.94\%) and Evidence (89.90\%) scores. This superior performance is likely due to the deeper learning capabilities and broader contextual understanding inherent in larger models. Variations in performance within this category also highlight the potential for specific architectural enhancements and targeted training strategies, as evidenced by the disparity between Llama2 and Llama3.

\textbf{Fine-tuning PLMs and LLMs}
Fine-tuning both PLMs and LLMs consistently produces better results than prompting methods. Fine-tuning, which involves specific adjustments to model weights for the task, enables the models to directly learn detailed and nuanced patterns within the training data. The effectiveness of fine-tuning is particularly evident in scenarios involving Gold Evidence, where the fine-tuned model can precisely assess the validity of claims based on key information.

\textbf{Performance with Gold Evidence versus Full Context} The use of Gold Evidence typically results in higher accuracy scores across models compared to when the Full Context is provided. Gold evidence, being directly relevant to the claims, allows models to focus their computational power on a smaller, more pertinent dataset, thereby reducing the noise associated with broader contexts. This targeted approach leads to more precise verifications but does not necessarily prepare models for real-world scenarios where they must extract relevant information from extensive, unstructured data.

\textbf{Prompting and Handling Full Context} Models designed to handle extensive and complex contexts, such as Gemini, benefit from prompting techniques that leverage pre-trained knowledge to interpret new data without extensive re-training. This approach enables efficient navigation and processing of large datasets, making it especially suitable for applications that require the processing of generalized information. However, despite its capability to manage broader data, prompting generally falls short of achieving the accuracy delivered by fine-tuning, particularly when detailed specificity and deep data understanding are necessary.

\textbf{Influence of Model Architecture and Size} The results consistently reveal that larger models such as XLM-R$_{large}$ and Gemma surpass their smaller counterparts in both context and evidence metrics. The enhanced performance of these models is attributed to their expanded capacity, which is essential for addressing the intricacies associated with verifying claims. Equipped with extensive neural networks and deeper layers, these models possess greater computational power, enabling them to effectively model complex relationships and dependencies in the data. This allows for more effective information extraction and synthesis, providing a significant advantage in fact-checking tasks.

\subsection{Analysis and Discussion}
\paragraph{How does the Evidences Retrieval help?}
Our analysis of retrieval models in fact-checking sheds light on the operational dynamics of SBERT \cite{reimers-2019-sentence-bert}, BM25 \cite{robertson2009probabilistic}, and their hybrid configurations under various conditions, with a focus on how well these models understand the semantic complexities of language processing (see Figure \ref{fig:evidences_ir}). The choice of these models for a more detailed evaluation is based on their superior performance across experiments, as discussed in Section \ref{main_results}.


A deeper dive into the results reveals that increasing the number of top-K retrieved evidences universally benefits all models by expanding the pool of potentially relevant information. However, the relationship between the number of documents retrieved (K) and the improvement in F1-score is not linear and varies significantly between different models and configurations. SBERT, in particular, shows a strong positive correlation between increased K and performance gains, indicating its effective use of broader contextual data.

Interestingly, performance improvements begin to plateau at higher K values in certain configurations, including Gemma within the SBERT model, suggesting an optimal K threshold of 5. This threshold represents the balance point where the benefits of additional document retrieval begin to decline relative to the computational costs. This insight is crucial for optimizing retrieval systems, emphasizing the need to balance data comprehensiveness with resource efficiency. 

Furthermore, the distinct behavior of configurations like Gemini 1.5 Flash under SBERT, which scales effectively with an increase in K, underscores the potential for tailored approaches based on specific system capabilities and task requirements. Such adaptability is crucial in cases where the volume and variety of information vary dramatically.

\begin{table*}[ht]
\centering
\resizebox{.95\textwidth}{!}{%
\begin{tabular}{lcccccccccccc}
\hline
\multicolumn{1}{l|}{\multirow{2}{*}{}} & \multicolumn{4}{c|}{\textbf{Single-evidence}} & \multicolumn{4}{c|}{\textbf{Multiple-evidence}} & \multicolumn{4}{c}{\textbf{Overall}} \\ \cline{2-13} 
\multicolumn{1}{l|}{} & \textbf{Avg. F1} & \textbf{Support} & \textbf{Refute} & \multicolumn{1}{c|}{\textbf{NEI}} & \textbf{Avg. F1} & \textbf{Support} & \textbf{Refute} & \multicolumn{1}{c|}{\textbf{NEI}} & \textbf{Avg. F1} & \textbf{Support} & \textbf{Refute} & \textbf{NEI} \\ \hline
\multicolumn{13}{c}{\textbf{\textit{Fine-tuning PLMs}}} \\ \hline
\multicolumn{1}{l|}{PhoBERT$_{base}$} & 69.79 & 71.04 & 65.53 & \multicolumn{1}{c|}{72.80} & 64.92 & 66.89 & 67.10 & \multicolumn{1}{c|}{60.77} & 68.47 & 69.75 & 69.75 & 69.75 \\
\multicolumn{1}{l|}{\textbf{PhoBERT$_{large}$}} & 75.01 & 76.27 & 70.81 & \multicolumn{1}{c|}{77.95} & 62.72 & 64.44 & 67.12 & \multicolumn{1}{c|}{56.58} & 71.47 & 72.64 & 69.59 & 72.18 \\
\multicolumn{1}{l|}{ViBERT} & 56.42 & 61.38 & 46.31 & \multicolumn{1}{c|}{61.59} & 58.11 & 63.64 & 56.23 & \multicolumn{1}{c|}{54.46} & 57.16 & 62.08 & 49.57 & 59.81 \\ 
\multicolumn{1}{l|}{mBERT} & 71.92 & 70.45 & 67.17 & \multicolumn{1}{c|}{78.13} & 61.72 & 62.19 & 60.57 & \multicolumn{1}{c|}{62.41} & 68.97 & 67.80 & 65.03 & 74.08 \\
\multicolumn{1}{l|}{XLM-R$_{base}$} & 71.52 & 74.96 & 64.75 & \multicolumn{1}{c|}{74.86} & 67.46 & 67.35 & 69.49 & \multicolumn{1}{c|}{65.56} & 70.48 & 72.57 & 66.33 & 72.54 \\
\multicolumn{1}{l|}{\textbf{XLM-R$_{large}$}} & \underline{80.06} & \underline{ 81.38} & \underline{ 78.00} & \multicolumn{1}{c|}{\underline{ 80.80}} & \underline{ 74.97} & \underline{ 78.96} & \underline{ 77.82} & \multicolumn{1}{c|}{\underline{ 68.14}} & \underline{ 78.75} & \underline{ 80.60} & \underline{ 77.94} & \underline{ 77.72} \\ \hline
\multicolumn{13}{c}{\textbf{\textit{Fine-tuning LLMs}}} \\ \hline
\multicolumn{1}{l|}{\textbf{Gemma}} & \underline{ \textbf{83.99}} & 84.77 & 82.33 & \multicolumn{1}{c|}{\underline{ \textbf{84.87}}} & \underline{ \textbf{79.52}} & 81.71 & \underline{ \textbf{81.96}} & \multicolumn{1}{c|}{\underline{ \textbf{74.88}}} & \underline{ \textbf{82.85}} & 83.75 & \underline{ \textbf{82.21}} & \underline{ \textbf{82.59}} \\
\multicolumn{1}{l|}{Mistral} & 83.62 & 85.26 & \underline{ \textbf{83.01}} & \multicolumn{1}{c|}{82.60} & 77.35 & \underline{ \textbf{82.99}} & 78.11 & \multicolumn{1}{c|}{70.94} & 81.89 & 84.52 & 81.41 & 79.75 \\
\multicolumn{1}{l|}{Llama2} & 38.99 & 38.77 & 33.62 & \multicolumn{1}{c|}{44.59} & 38.05 & 45.02 & 33.09 & \multicolumn{1}{c|}{36.04} & 38.81 & 40.73 & 33.45 & 42.27 \\
\multicolumn{1}{l|}{Llama3} & 83.45 & \underline{ \textbf{85.88}} & 80.64 & \multicolumn{1}{c|}{83.82} & 75.02 & 82.01 & 77.51 & \multicolumn{1}{c|}{65.55} & 81.18 & \underline{ \textbf{84.59}} & 79.65 & 79.29 \\ \hline
\multicolumn{13}{c}{\textbf{\textit{Prompting LLMs}}} \\ \hline
\multicolumn{1}{l|}{\textbf{Gemini}} & \underline{ 73.96} & \underline{ 80.85} & \underline{ 71.58} & \multicolumn{1}{c|}{\underline{ 69.46}} & \underline{ 75.33} & \underline{ 80.55} & \underline{ 72.70} & \multicolumn{1}{c|}{\underline{ 72.75}} & \underline{ 69.96} & \underline{ 81.46} & \underline{ 69.29} & \underline{ 59.13} \\
\multicolumn{1}{l|}{Gemma} & 49.53 & 53.33 & 54.16 & \multicolumn{1}{c|}{41.09} & 52.08 & 55.67 & 56.19 & \multicolumn{1}{c|}{46.55} & 49.76 & 61.26 & 52.31 & 35.71 \\
\multicolumn{1}{l|}{Mistral} & 51.54 & 68.79 & 53.38 & \multicolumn{1}{c|}{32.45} & 53.97 & 68.20 & 53.01 & \multicolumn{1}{c|}{40.69} & 49.99 & 68.06 & 50.14 & 31.78 \\
\multicolumn{1}{l|}{Llama2} & 36.12 & 65.43 & 11.95 & \multicolumn{1}{c|}{30.99} & 43.58 & 64.08 & \multicolumn{1}{r}{30.83} & \multicolumn{1}{c|}{35.83} & 33.16 & 67.14 & 19.68 & 22.66 \\
\multicolumn{1}{l|}{Llama3} & 48.65 & 61.16 & 50.41 & \multicolumn{1}{c|}{34.37} & 52.31 & 55.31 & 59.56 & \multicolumn{1}{c|}{42.06} & 43.71 & 48.21 & 46.25 & 36.67 \\ \hline
\end{tabular}%
}
\caption{Performance comparison of language models across Single and Multiple evidence scenarios.}
\label{tab:sing_multi}
\end{table*}

\paragraph{How Multi-evidence Impacts Model Reasoning?}
The comparative performance of language models shows significant variations, particularly when comparing their ability to handle single-evidence versus multiple-evidence inputs, as depicted in Table \ref{tab:sing_multi}. Gemma stands out for its robust capability in both of scenarios, benefitting significantly from training on a diverse, multilingual dataset. This extensive training enhances its adaptability and accuracy by enabling it to effectively manage complex contexts. Additionally, Gemma excels in data sufficiency assessments, effectively classifying the Not Enough Information (NEI) category across different scenarios, which is crucial for ensuring the reliability of fact-checking systems and preventing misinformation.

In single-evidence scenarios, the simplicity of the data allows models such as Llama3 to achieve higher accuracy. This straightforwardness typically presents less ambiguity, enabling the models to apply their verification capabilities more effectively. However, when multiple evidence sources are introduced, the added complexity significantly challenges all models. The noticeable decline in performance metrics in these scenarios highlights a gap in the ability of models to synthesize and integrate information from various sources, revealing a critical area for future enhancements.


\subsection{Qualitative Error Analysis}
Based on the macro F1 scores, we selected the Gemma model as our baseline to perform a detailed error analysis. As illustrated in Figure \ref{fig:error_analysis} and further detailed in Appendix K, we evaluated 100 random incorrect predictions from the development set to identify and categorize error types.

\begin{figure}[ht]
    \centering
    \includegraphics[width=0.48\textwidth]{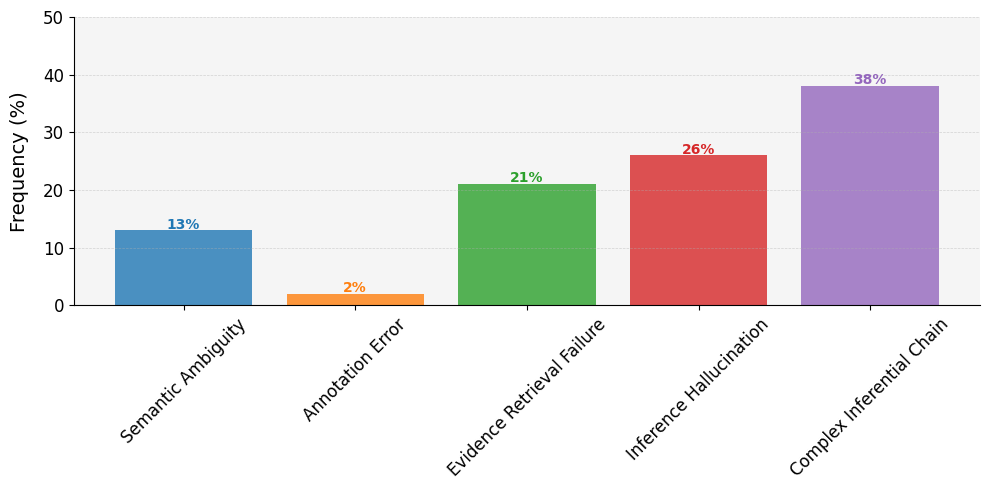}
    \caption{Distributions of errors.}
    \label{fig:error_analysis} 
\end{figure}

The analysis revealed significant challenges in handling Semantic Ambiguity and Complex Inferential Chains, both of which are pivotal for refining NLP technologies. Semantic Ambiguity issues particularly highlight the necessity for context-aware processing \cite{wang-etal-2022-pan,wu-etal-2023-ambiguous}. By integrating transformer-based models, the ability of the Gemma model to interpret complex linguistic contexts could be substantially improved, enhancing its accuracy in environments where nuance is critical.

Moreover, the frequent errors associated with Complex Inferential Chains expose the limitations of the model in synthesizing and reasoning across diverse informational inputs. The adoption of memory networks and knowledge graphs could markedly improve its capacity to process and link extended data sequences, thereby enhancing its reasoning and inference capabilities \cite{kim-etal-2023-factkg,pan-etal-2023-fact}.

\subsection{Human Performance}

Table \ref{table:test_200} presents an evaluation of fine-tuned models, offering crucial insights into their varied performances in the Support, Refute, and NEI compared to human performance. Models such as Gemma and Llama3 demonstrate strong capabilities in the Support and NEI categories, indicating their robustness in handling both direct and ambiguous information. However, their performance declines in the Refute category, highlighting a critical gap in the ability of AI to effectively process and analyze contradictory information.

\begin{table}[ht]
\centering
\resizebox{.95\linewidth}{!}{%
\begin{tabular}{lcccc}
\hline
\multicolumn{1}{c}{\textbf{Model}} & \textbf{F1 score} & \textbf{Support} & \textbf{Refute} & \textbf{NEI} \\ \hline
\multicolumn{5}{c}{\textbf{Fine-tuning PLMs}} \\ \hline
\multicolumn{1}{l}{PhoBERT$_{base}$} & 71.29 & 75.19 & 63.89 & 74.80 \\
\multicolumn{1}{l}{PhoBERT$_{large}$} & 73.08 & 79.70 & 62.30 & 77.24 \\
\multicolumn{1}{l}{ViBERT} & 55.66 & 68.70 & 48.28 & 50.00 \\
\multicolumn{1}{l}{mBERT} & 66.94 & 71.79 & 61.84 & 67.18 \\
\multicolumn{1}{l}{XLM-R$_{base}$} & 66.33 & 71.64 & 64.97 & 62.39 \\
\multicolumn{1}{l}{XLM-R$_{large}$} & 74.95 & 76.47 & 73.02 & 75.36 \\ \hline
\multicolumn{5}{c}{\textbf{Fine-tuning LLMs}} \\ \hline
\multicolumn{1}{l}{Gemma} & 83.95 & \colorbox{gray!30}{91.73} & 77.52 & \colorbox{gray!30}{82.61} \\
\multicolumn{1}{l}{Mistral} & 66.61 & 77.46 & 62.69 & 59.68 \\
\multicolumn{1}{l}{Llama2} & 46.10 & 50.45 & 40.94 & 46.91 \\
\multicolumn{1}{l}{Llama3} & 84.24 & \colorbox{gray!30}{91.97} & 77.05 & \colorbox{gray!30}{83.69} \\ \hline
\multicolumn{5}{c}{\textbf{Human Evaluating}} \\ \hline
\multicolumn{1}{l}{Human} & \textbf{84.93} & \textbf{81.25} & \textbf{80.95} & \textbf{82.38} \\ \hline
\end{tabular}%
}
\caption{Evaluation results of human performance compared to the models on the test set of 200 samples. Models that outperform human evaluators are marked in gray.}
\label{table:test_200}
\end{table}

This pattern is not isolated but is evident across various models, suggesting that current AI architectures and training paradigms may lack the sophisticated reasoning required to handle complex linguistic challenges that humans manage more adeptly. The comparative underperformance of AI in the Refute category underscores the need for integrating deeper contextual understanding and advanced reasoning mechanisms into AI systems to better mimic human cognitive abilities in processing contradictions and complex arguments.

\section{Conclusion \& Future Work}\label{conclusion}
The development of the ViFactCheck dataset marks a transformative advancement in fact-checking for Vietnamese. This dataset comprises 7,232 entries across 12 topics, providing a substantial resource to assess various SOTA baseline models. Our work demonstrates the potential of using advanced language models, fine-tuned on this dataset, to achieve high levels of accuracy, as evidenced by a macro F1 score of 89.90\%. This validates the efficacy of our dataset and methodologies in a real-world context, setting a new benchmark for fact-checking performance in low-resource languages. The challenges identified through our in-depth analysis, such as semantic ambiguity and evidence retrieval failures, not only underscore the complexity of fact-checking in such environments but also pave the way for targeted improvements.


Future research will focus on addressing the identified challenges to further enhance model performance. Efforts will include refining semantic understanding and evidence retrieval capabilities to handle ambiguous and complex datasets more effectively \cite{wang-etal-2022-pan,wu-etal-2023-ambiguous}. In addition, we plan to develop methods to mitigate inference hallucinations and improve reasoning across complex inferential chains \cite{kim-etal-2023-factkg,pan-etal-2023-fact}. Expanding the dataset to incorporate a wider range of misinformation types and correcting labeling errors will also be crucial \cite{gupta2021xfact,augenstein-etal-2019-multifc}.




\section*{Acknowledgments}
This research was supported by The VNUHCM-University of Information Technology's Scientific Research Support Fund of the VNUHCM University of Information Technology.

\bibliography{aaai25}

\appendix

\section{Task Definition} \label{sec:task_def}
Fact-checking can be defined as the task of automatically assessing the veracity of a given claim based on available evidence. This task typically involves two main steps: \textbf{1) Evidence Retrieval (Fact Extraction):} This step aims to find relevant evidence from a given corpus to support or refute the claim. \textbf{2) Claim Verification (Fact Verification):} This step determines the truthfulness of the claim based on the retrieved evidence.

The proposed system is designed to assign labels to the claims, categorizing them as follows:

\begin{itemize}
    \item \textbf{Support:} The claim is confirmed to be correct according to the available evidence.
    \item \textbf{Refute:} The claim is determined to be inaccurate compared to the available evidence.
    \item \textbf{Not Enough Info (NEI):} The claim is not sufficiently supported by the evidence within the corresponding news, making it impossible to definitively verify or refute.
\end{itemize}

The goal of fact-checking systems is to assist human fact-checkers in their efforts to combat the spread of disinformation and false news. These systems provide automated tools that assess the credibility of claims in various sources, including news articles, social media, and political speeches.

\section{Data Collection Source} \label{sec:data_sources}
The data for this study was collected from reliable, government-licensed online newspaper websites in Vietnam, which boast significant visitor counts and provide up-to-date news. The sources included Bao Chinh Phu, VnExpress, Dan Tri, Nguoi Lao Dong, Tuoi Tre, Tin Tuc, Phap Luat HCM, Thanh Nien, and Tien Phong. To extract news articles from these online newspaper websites, we utilized two Python libraries, BeautifulSoup and Selenium. These libraries are well-known for their robust capabilities in extracting data from websites.

\begin{table}[ht]
\begin{adjustbox}{width=0.5\textwidth}
\begin{tblr}{
  hline{1-2,10} = {-}{},
}
\textbf{Website} & \textbf{Organization} & \textbf{URL}\\
Bao Chinh Phu & Government of Vietnam & \url{https://baochinhphu.vn}\\
VnExpress & MOST Vietnam & \url{https://vnexpress.net}\\
Dan Tri & MOLISA Vietnam & \url{https://dantri.com.vn}\\
Nguoi Lao Dong & HCM City Committee & \url{https://nld.com.vn}\\
Tuoi Tre & HCM Communist Youth Union  & \url{https://tuoitre.vn}\\
Tin Tuc & Vietnam News Agency & \url{https://baotintuc.vn}\\
Phap Luat HCM & HCM City People's Committee & \url{https://plo.vn}\\
Thanh Nien & Vietnam Youth Union & \url{https://thanhnien.vn}
\end{tblr}
\end{adjustbox}
\caption{Details of the sources and organizations of the online news sites in the ViFactCheck dataset.}
\end{table}

\section{Data Annotation Tool and Guideline} \label{sec:guide_line}
\begin{figure}[ht]
    \centering
    \includegraphics[width=\linewidth]{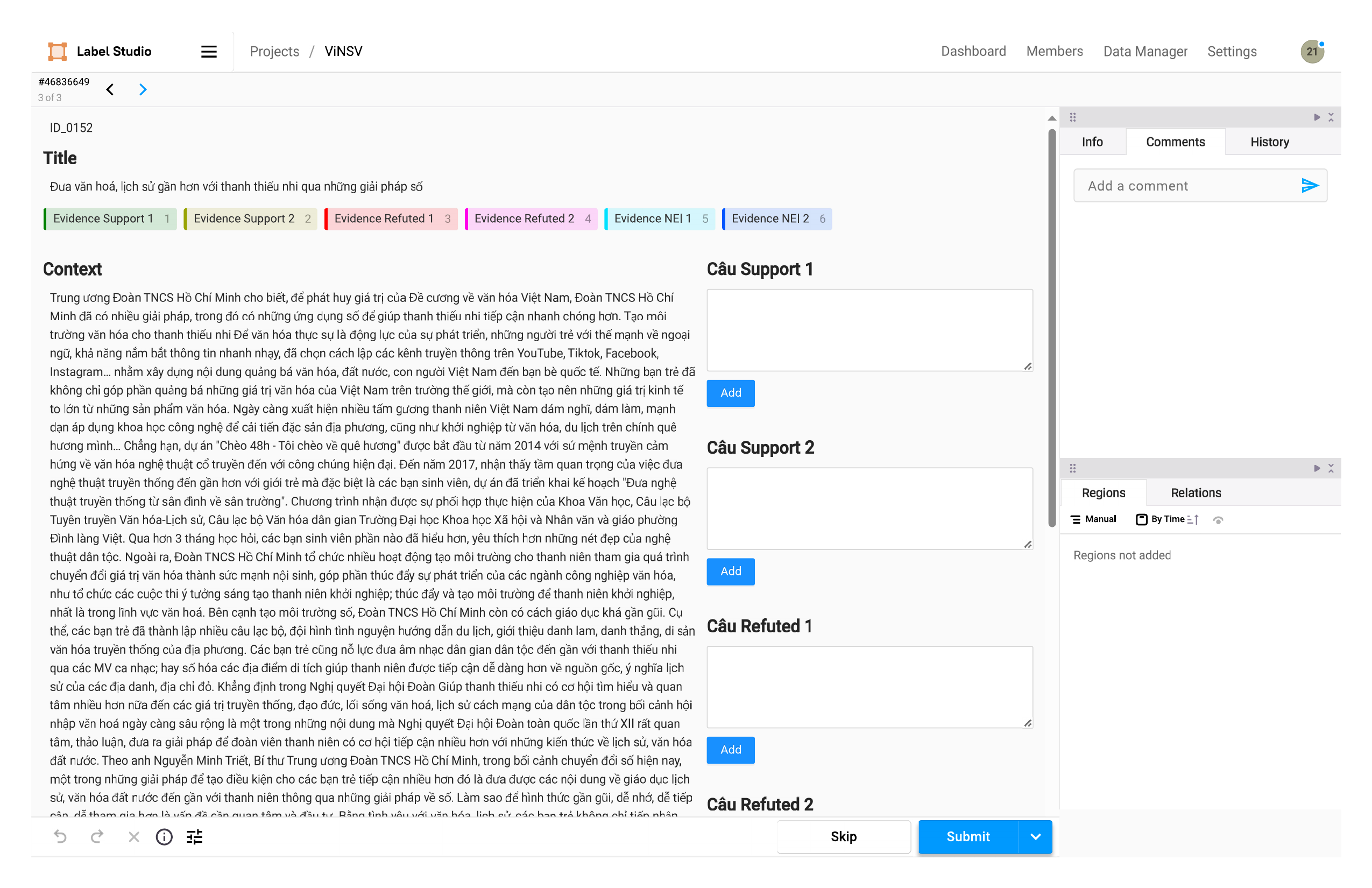}
    \caption{Label Studio UI for our annotation task.}
    \label{fig:annotation_tool} 
\end{figure}

\paragraph{Data annotation tool} During the annotation phases of the dataset, we utilized Label Studio, an open-source platform that provides an intuitive interface and supports various labeling tasks across different types of data. Our annotation interface is shown in Figure \ref{fig:annotation_tool}.

\paragraph{Data annotation guideline}
Comprehensive guidelines were provided to the annotators to ensure a cohesive and systematic approach:

\textbf{(1)} The annotation process required the generation of six claim pairs for each article in the dataset, resulting in two pairs for each designated label: \textbf{Support, Refute, and NEI (Not Enough Information).}
\textbf{(2)} For the \textbf{Support} and \textbf{Refute} labels, annotations were grounded in the intrinsic information and contextual evidence derived directly from the corresponding news articles. The \textbf{NEI} label required a more nuanced approach, involving the addition of external information and context, which could either align with or deviate from the truth.
\textbf{(3)} The generated claims must adhere to certain rules: paraphrasing sentences from the article, inferring claims by combining multiple pieces of information, and meticulously avoiding spelling and abbreviation errors that could compromise the quality of the dataset. 
\textbf{(4)} To enrich the dataset with diverse perspectives and challenges, annotators were encouraged to leverage their broad vocabulary and skilled sentence-writing techniques, thereby introducing valuable nuances into the annotations.

\section{Human Recruitment}\label{sec:anno_rec}
\paragraph{Annotation Recruitment}
In our study, we recruited seven university students as annotators, all native Vietnamese speakers aged 20 to 22, representing a diverse range of academic disciplines. These disciplines included social sciences, natural sciences, and Vietnamese studies. Their selection was based on exceptional linguistic skills, demonstrated by high scores in Vietnamese literature exams, and a deep familiarity with various media platforms, ensuring annotations that accurately reflect current linguistic trends.

To ensure the reliability and accuracy of our annotation process, we engaged two linguistic experts with a solid background in Vietnamese grammar and syntax. These experts, also native speakers, were tasked with developing the guidelines and continuously monitoring the annotation process. Their expertise is grounded in extensive academic achievements and a critical ability to evaluate news content across various media platforms, adding a significant layer of scientific rigor and depth to our data creation methodology.

\definecolor{JapaneseLaurel}{rgb}{0,0.501,0}
\definecolor{WebOrange}{rgb}{1,0.647,0}

\begin{table*}[]
\centering
\begin{tabular}{@{}lp{14.7cm}@{}}
\toprule
\textbf{Context} & TPO - \textcolor{red}{Tổng Công ty Cảng Hàng không Việt Nam (ACV) vừa chính thức gia hạn thời gian mời thầu thêm 1 tháng}, kéo dài thời gian thực hiện gói thầu thi công nhà ga sân bay Long Thành từ 33 tháng lên 39 tháng. \textcolor{red}{Như vậy, ``siêu sân bay'' Long Thành sẽ chỉ có thể đưa vào khai thác từ năm 2026 thay vì mục tiêu năm 2025 như trước đó}. Tin từ ACV cho hay, \textcolor{JapaneseLaurel}{đơn vị chính thức điều chỉnh kế hoạch và hồ sơ mời thầu gói thầu thi công xây dựng và lắp đặt thiết bị nhà ga hành khách sân bay Long Thành giai đoạn 1} (do ACV làm chủ đầu tư). Cụ thể, \textcolor{JapaneseLaurel}{thời gian mời thầu} được gia hạn thêm 1 tháng, \textcolor{JapaneseLaurel}{kéo dài tới sáng ngày 28/4, thay vì tới ngày 28/3 như trước đó.} ... \textcolor{WebOrange}{Gói thầu thi công nhà ga hành khách sân bay Long Thành trị giá hơn 35 nghìn tỷ đồng} do ACV làm chủ đầu tư. \textcolor{WebOrange}{Đây là gói thầu lớn nhất dự án sân bay Long Thành}... 

(\textbf{English:} TPO - \textcolor{red}{Vietnam Airport Corporation (ACV) has officially extended the bidding period by an additional month}, prolonging the implementation time for the construction contract of the Long Thanh Airport passenger terminal from 33 to 39 months. \textcolor{red}{Consequently, the ``mega airport'' Long Thanh will only be operational by 2026 instead of the previous target of 2025}. According to ACV, \textcolor{JapaneseLaurel}{the organization has formally adjusted the plan and tender documents for the construction and installation of the passenger terminal at Long Thanh Airport Phase 1} (with ACV as the main investor). Specifically, the \textcolor{JapaneseLaurel}{bidding period} has been extended by one month, \textcolor{JapaneseLaurel}{now ending on the morning of April 28, instead of the previous deadline of March 28}. ... \textcolor{WebOrange}{The construction contract for Long Thanh Airport's passenger terminal, valued at over 35 trillion VND} is being managed by ACV. \textcolor{WebOrange}{This is the largest contract within the Long Thanh Airport project.})  
\\ \\ \hdashline \\
 \textcolor{JapaneseLaurel}{\textbf{Support}} & Việc nhà thầu thi công xây dựng và lắp đặt thiết bị nhà ga hành khách sân bay Long Thành giai đoạn 1 bị điều chỉnh, thời gian bị kéo dài tới sáng ngày 28/4 thay vì tới ngày 28/3 như dự kiến.
 
\textbf{English:} \textit{The construction and installation contract for the Long Thanh Airport Phase 1 passenger terminal has been adjusted, with the timeline extended to the morning of April 28 instead of the originally anticipated March 28.}
\\
 
 \textcolor{red}{\textbf{Refute}} & Tổng Công ty Cảng Hàng không Việt Nam (ACV) vừa gia hạn thời gian mời thầu thêm thời gian 2 tháng, tức ``siêu sân bay'' Long Thành sẽ chỉ có thể đưa vào sử dụng từ năm 2026 thay vì năm 2025 như dự kiến ban đầu.
 
\textbf{English:} \textit{Vietnam Airport Corporation (ACV) has recently extended the bidding period by an additional 2 months, meaning that the ``mega airport'' Long Thanh will only be operational by 2026 instead of the originally planned year 2025.}
\\
  
 \textcolor{WebOrange}{\textbf{NEI}} & Gói thầu lớn nhất dự án sân bay Long Thành là gói thầu thi công nhà ga hành khách với trị giá hơn 35 nghìn tỷ đồng, được tài trợ bởi công ty Hàn Quốc.  
 
 \textbf{English:} \textit{The largest contract within the Long Thanh Airport project is the construction of the passenger terminal, valued at over 35 trillion VND, and it is sponsored by a South Korean company.}
 \\ \bottomrule
\end{tabular}
\caption{Typical samples from the \textbf{ViFactCheck} dataset with three labels \textcolor{JapaneseLaurel}{\textbf{Support}}, \textcolor{red}{\textbf{Refute}}, and \textcolor{WebOrange}{\textbf{NEI}}. The highlighted words is the evidence of the claim.}
\label{table:dataexample}
\end{table*}

\paragraph{Human Evaluation Recruitment}
To evaluate human performance in the fact-checking process, we engaged three native Vietnamese-speaking students who had no prior exposure to the task of fact-checking. They were tasked with annotating a representative subset consisting of 200 samples. Comprehensive instructions were provided to ensure their understanding of the task, including clarifications on the significance of each label and additional information to assist them in determining the appropriate labels for each sample. The final label for each claim was determined through a majority consensus among the assessors.

\section{Data Examples} \label{sec:data_sample}
The ViFactCheck dataset includes various examples of written claims, as illustrated in Table \ref{table:dataexample}. To create a challenging and realistic context, annotators were tasked with generating claims based on multiple evidences, which are highlighted within the textual context provided. This approach not only enhances the complexity and challenge of the annotation task but also contributes significantly to the reliability and practical value of the dataset for fact-checking tasks in the Vietnamese language. By ensuring that claims are grounded in verifiable pieces of evidence, the dataset fosters a robust environment for training and evaluating language models specifically tailored to the nuances of fact verification.

\section{Human-Generated Rules} \label{sec:rule base}
In the ViFactCheck datasets, annotators were encouraged to leverage their broad vocabulary and skilled sentence-writing techniques, thus introducing valuable nuances into the annotations. The basic rules for the use of the generation by annotators are summarized in Table \ref{table:rule}.

\begin{table}[ht]
\centering
\begin{adjustbox}{width=0.49\textwidth}
\begin{tblr}{
    colspec={Q[l] Q[] Q[r]},
    cell{1}{2} = {c},
    cell{2}{1} = {r=4}{},
    cell{6}{1} = {r=6}{},
    cell{12}{1} = {r=2}{},
    hline{1,14} = {-}{0.08em},
    hline{2,6,12} = {-}{},
    hline{3-5,7-11,13} = {2-3}{dashed},
}
                  & \textbf{Rules}                                        & \textbf{Ratio (\%)} \\
\begin{sideways}\textbf{Support}\end{sideways} & Restructuring the evidences                          & 73.68               \\
                  & Eliminating or adding words & 44.21               \\
                  & Substituting numbers, time, or mathematical inferences & 7.34                \\
                  & Altering the word order in a sentence & 8.42                \\
\begin{sideways}\textbf{Refute}\end{sideways} & Employing Negation                                   & 8.16                \\
                  & Replacing Words with Antonyms & 17.35               \\
                  & Misrepresenting quantity & 22.45               \\
                  & Misrepresenting Temporal Logic & 16.37               \\
                  & Misinterpreting Entity Relationships & 5.11                \\
                  & Misjudging Event Dynamics & 47.96               \\
\begin{sideways}\textbf{NEI}\end{sideways} & Inferring sentences with unspecified information                     & 90.20               \\
                  &      Utilizing external knowledge      & 10.78               
\end{tblr}
\end{adjustbox}
\caption{Approaches and rules for generating claims by humans in the ViFactCheck dataset. Note that a claim could involve multiple rules}
\label{table:rule}
\end{table}

\begin{figure*}[ht]
    \centering
    \includegraphics[width=\textwidth]{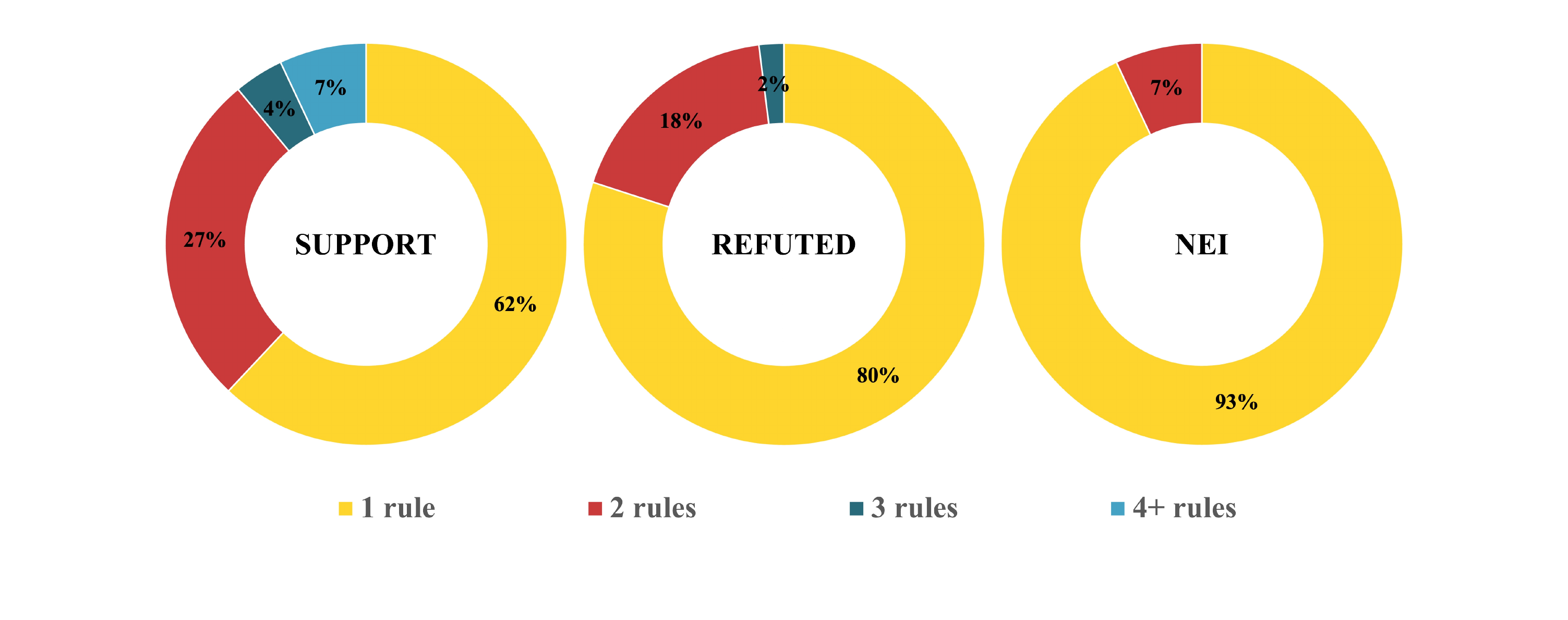}
    \caption{The ratio of combining different rules to create claims in ViFactCheck.}
    \label{fig:RULES_Ratio} 
\end{figure*}

Annotators are required to follow guidelines to create diverse and challenging data. The distribution of data-generating rule usage for claims related to Support, Refute, and Not Enough Information (NEI) is shown in Figure \ref{fig:RULES_Ratio}. To understand how annotators behave in creating ViFactCheck, we analyzed the number of rules used to generate claims. We randomly selected 100 context-claim pairs for Support, Refute, and NEI categories.

The primary trend in this dataset reveals an obvious bias towards using 1-2 rules, reflecting a standardized annotation process. However, some annotators deviated from this trend, opting for four or more rules, demonstrating an awareness of the complexity and diversity of data. This underscores the importance of judiciously combining rules for reliable and accurate annotation.

The use of multiple rules presents challenges for language model development, introducing complexity into inference and decision-making processes dependent on rule combinations. However, it also offers an opportunity to improve more adaptable language models, ensuring greater accuracy in making inferences.

\section{Additional Dataset Analysis}\label{sec:add_analysis}
\paragraph{Dataset basic statistic.} The ViFactCheck dataset contains 7,232 samples divided into three subsets: training, development, and test with a ratio of 7:1:2. The basic statistics of the three subsets are shown in Table \ref{table:Statistic}. We observed that the average length of a context in the dataset is approximately 700 words, with the longest context extending to 3,602 words. Such richness in context proves highly beneficial for models with large parameter sets, such as Gemma, as they can effectively capture the maximum features of the data. On average, each claim sentence contains about 36 words, with the longest reaching 165 words.

\begin{table}[ht]
\centering
\begin{adjustbox}{width=0.48\textwidth}
\begin{tblr}{
  row{1} = {c},
  cell{2}{1} = {r=5}{c},
  cell{2-11}{3-5} = {r},
  cell{7}{1} = {r=5}{c},
  vline{3-5} = {1}{-}{},
  vline{3} = {2}{-}{},
  hline{1} = {-}{0.08em},
  hline{2,7,12} = {-}{},
  hline{3-6,8-11} = {2-5}{dashed},
}
 &  & \textbf{Training} & \textbf{Development} & \textbf{Test}\\
\begin{sideways}\textbf{Context}\end{sideways} & Total samples & 1035 & 496 & 758\\
 & Avg length & 693.2 & 670.2 & 690.5\\
 & Max length & 3602 & 2534 & 3602\\
 & Min length & 71 & 71 & 71\\
 & Total vocab size & 25,382 & 16,522 & 21,263\\
\begin{sideways}\textbf{claim}\end{sideways} & Total samples & 5062 & 723 & 1447\\
 & Avg length & 35.9 & 35.6 & 35.8\\
 & Max length & 165 & 145 & 135\\
 & Min length & 7 & 10 & 7\\
 & Total vocab size & 12,189 & 4,555 & 6,711
\end{tblr}
\end{adjustbox}
\caption{Basic statistic of ViFactCheck dataset. The size and length of the vocab are computed at word level.}
\label{table:Statistic}
\end{table}

\begin{figure}[ht]
    \centering
    \includegraphics[width=\linewidth]{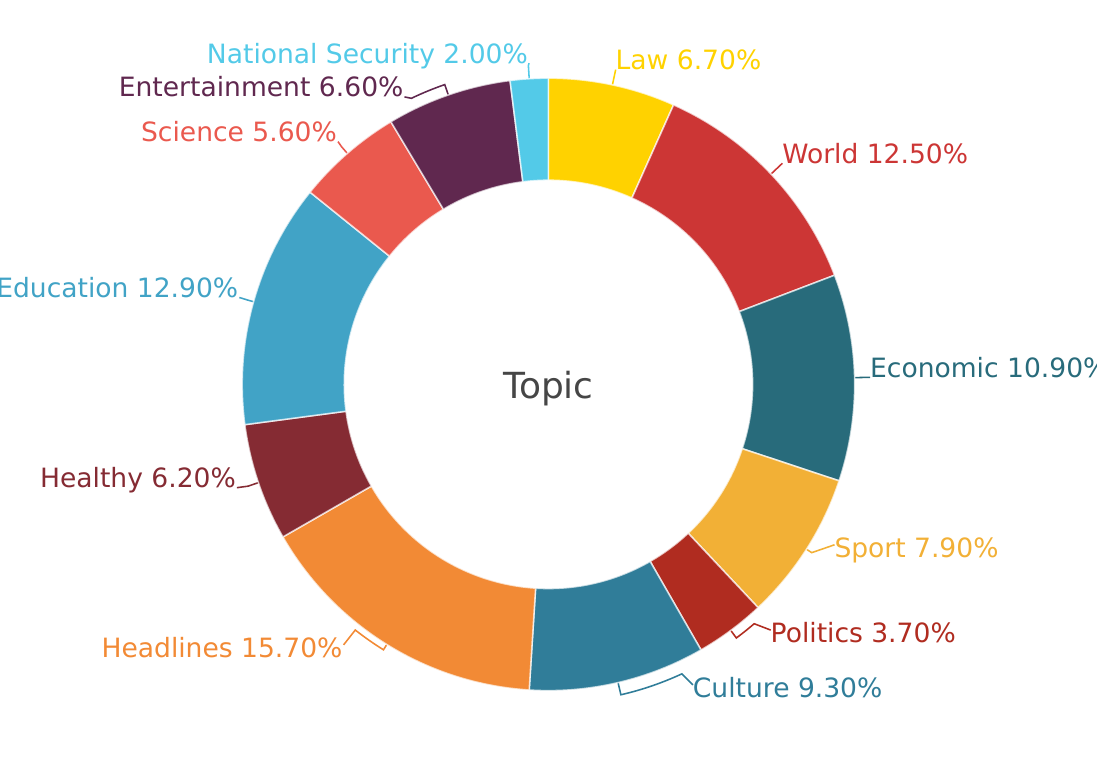}
    \caption{Topic distribution on ViFactCheck dataset.}
    \label{fig:statistic-chart} 
\end{figure}

\paragraph{Topic Distribution Analysis}\label{sec:topic}
The ViFactCheck dataset covers 12 popular topics frequently found in Vietnamese news, which are often subjected to misinformation. These topics, summarized in Figure \ref{fig:statistic-chart}, include ``Headlines'', ``World'', ``Education'', and ``Economics'', among others. ``Headlines'', covering updates on social issues and events, appears most frequently, demonstrating a significant presence in the dataset. The other notable topics, ``World", ``Education'', and ``Economics'', contribute 12.4\%, 12.9\%, and 10.9\% respectively. In contrast, ``National Security'' accounts for the lowest percentage at 2.0\%. This lower representation is attributed to the relatively few news on this topic in real life. Despite its smaller volume, due to the critical need for accuracy in information pertaining to national security, a concerted effort was made to include news related to this topic.

\begin{figure}[ht]
    \centering
    \includegraphics[width=\linewidth]{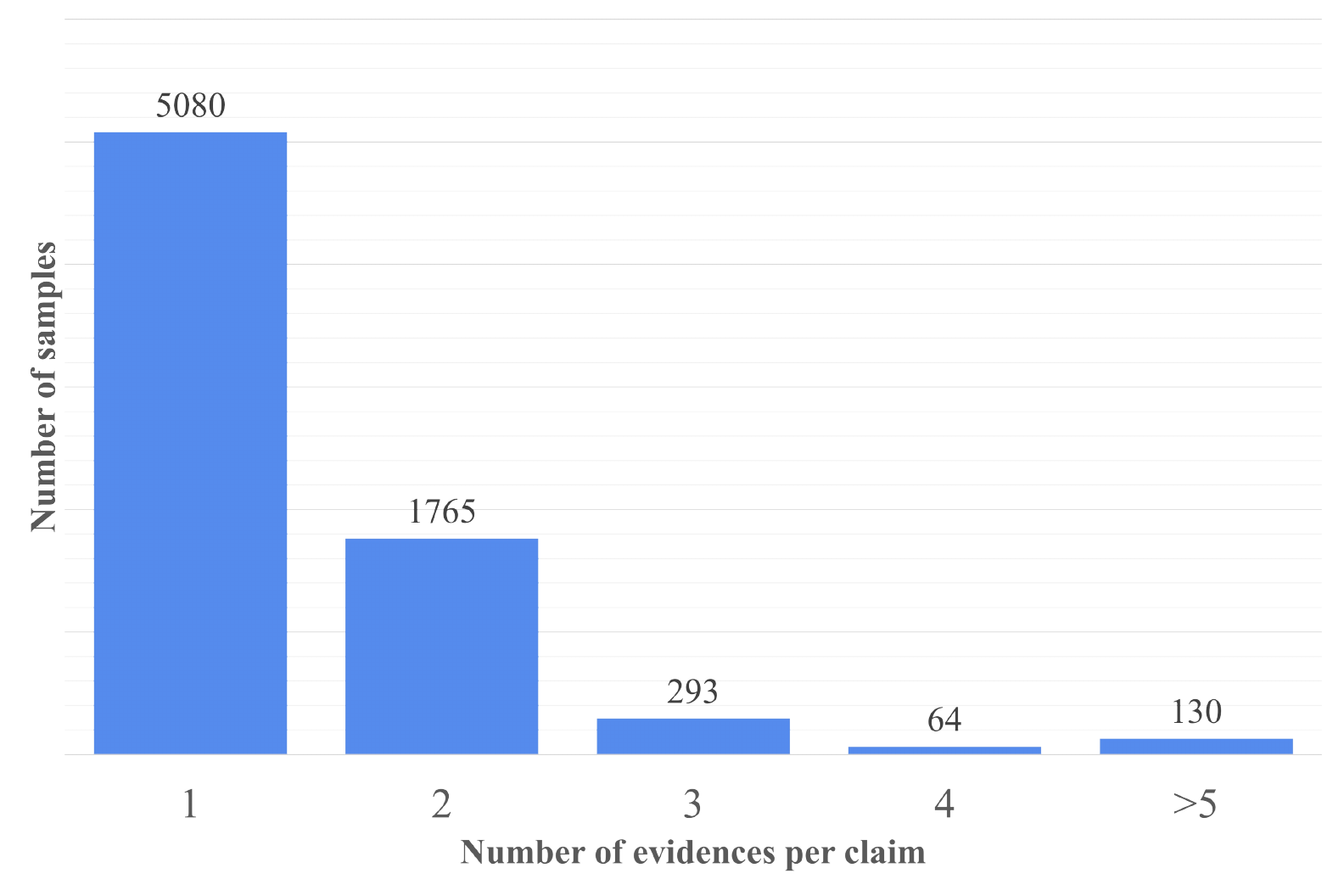}
    \caption{The distribution of single and multiple evidences samples in the ViFactCheck dataset.}
    \label{fig:evidence_ratio}
\end{figure}

\paragraph{Evidence Distribution Analysis}
Figure \ref{fig:evidence_ratio} from the ViFactCheck dataset shows the distribution of samples with varying numbers of evidence per claim. Single evidence refers to using only one piece of information to verify a claim, while multi-evidence involves integrating findings from multiple sources, necessitating advanced analytical skills to synthesize and validate information. 

The distribution reveals a predominant reliance on single pieces of evidence, where claims are supported by one source, reflecting simpler verification tasks. Multi-evidence scenarios, where claims are substantiated by two or more sources, demonstrate a steep decline to 1,765 and 293 samples for two and three evidences, respectively. This indicates the increasing complexity and computational demand of integrating diverse evidences. Notably, the rise to 130 samples for claims with more than five evidences suggests some scenarios necessitate extensive, complex reasoning, highlighting the capability of the dataset to train models for robust, multifaceted fact verification.

\section{Details about the Baselines}
\label{sec:baseline}
\subsection{Pre-trained Language Models}
Based on the significant performance of transformer-based in prior fact-checking tasks \cite{thorne-etal-2018-fever,hu2022chef,norregaard-derczynski-2021-danfever}, we employ pre-trained language models, specifically BERT \cite{devlin2019bert} and RoBERTa \cite{liu2019roberta} architectures, for the fact-checking task. This study includes four models, comprising two multilingual models and two monolingual models (the details of each model are shown in Table \ref{tab:models_compare}).

\textbf{mBERT} \cite{devlin2019bert} is a transformer-based model trained on an extensive corpus of 104 languages, including Vietnamese. Its linguistic versatility makes mBERT invaluable for fact-checking tasks. As a multilingual model, mBERT enables comprehensive analysis and serves as an excellent tool for ensuring the credibility of data within the Vietnamese fact-checking framework.

\textbf{Cross-lingual Language Model - RoBERTa (XLM-R)} \cite{conneau-etal-2020-unsupervised} is a transformer-based model trained on 100 languages. This vast linguistic scope means XLM-R can understand and compare information across different languages, an advantage for fact-checking that offers a broader context beyond the Vietnamese language. The ability of XLM-R to process information from multilingual sources or across language barriers is especially valuable when dealing with content that transcends linguistic boundaries.

\textbf{PhoBERT} \cite{nguyen2020phobert}, leveraging the powerful Transformer architecture of RoBERTa \cite{liu2019roberta}, exhibits a profound understanding of the nuances and context of the Vietnamese language. This linguistic precision is highly beneficial for the Vietnamese fact-checking dataset, as it can discern subtle language nuances that general models might overlook. With its focus on Vietnamese, PhoBERT delivers exceptional efficiency and accuracy when applied to a corpus of the same language, facilitating high-quality fact-checking within the Vietnamese context.
    
\textbf{ViBERT}, based on the BERT architecture and specifically designed for Vietnamese, was introduced by \citet{bui-etal-2020-improving}. Unlike mBERT, which is trained on a multi-language corpus, ViBERT is pre-trained on a substantial corpus of 10GB of uncompressed Vietnamese text, focusing solely on Vietnamese to achieve optimal performance.

By investigating the effectiveness of these BERT variants in Vietnamese fact-checking, we aim to enhance the field's ability to combat disinformation. The diversity of these models in terms of monolingual understanding, linguistic precision, and cross-lingual capabilities promises to make a contribution to the fact-checking landscape, advancing a more credible and precise information ecosystem.

\subsection{Fine-Tuning Large Language Models} \label{appendix:llms}
Recent advances in large language models (LLMs), which exhibit strong contextual understanding, have demonstrated their effectiveness in tasks such as contextual comprehension and reasoning, including fact-checking. Consequently, we employ several primary models that are suitable for low-resource configurations. Specifically, we utilize the open-source models Llama2 7B, Llama3 8B, Gemma 7B, and Mistral 7B.

\begin{table*}[ht]
\centering
\begin{adjustbox}{width=\textwidth}
\begin{tblr}{
  row{1} = {c},
  cell{2-12}{2} = {c},
  cell{2-12}{3} = {c},
  cell{2-11}{4-6} = {r},
  cell{12}{4-5} = {r},
  cell{12}{6} = {r},
  cell{7}{4} = {r},
  vline{2} = {-}{},
  vline{2} = {2}{-}{},
  hline{1} = {-}{0.08em},
  hline{2,5,8,13} = {-}{},
  hline{2} = {2}{-}{},
}
\textbf{Model} & \textbf{\#Layer} & \textbf{\#Head} & \textbf{\#Params} & \textbf{\#Vocab} & \textbf{\#MSL} & \textbf{Domain data} & \textbf{Language support}\\
PhoBERT$_{base}$ \cite{nguyen2020phobert} & 12 & 12 & 135M & 64K & 256 & ViWiki + ViNews & Vietnamese\\
PhoBERT$_{large}$ \cite{nguyen2020phobert} & 24 & 16 & 370M & 64K & 256 & ViWiki + ViNews & Vietnamese\\
ViBERT \cite{bui-etal-2020-improving}& 12 & 12 & - & 30K & 256 & Vietnamese News & Vietnamese\\
mBERT \cite{devlin2019bert} & 12 & 12 & 110M & 30K & 512 & Wikipedia + BookCorpus & Multilingual\\
XLM-R$_{base}$ \cite{conneau-etal-2020-unsupervised} & 12 & 12 & 270M & 250K & 512 & CommonCrawl & 100+ languages\\
XLM-R$_{large}$ \cite{conneau-etal-2020-unsupervised} & 24 & 16 & 550M & 250K & 512 & CommonCrawl & 100+ languages\\
Gemini \cite{reid2024gemini} & - & - & - & - & 1,048,576 & Mixed large datasets & Multilingual\\
Gemma \cite{team2024gemma} & 28 & 16 & 7B & 256K & 8,192 & Mixed large datasets & Multilingual\\
Mistral \cite{jiang2023mistral} & 32 & 32 & 7B & 32K & 32,768 & Mixed large datasets & Multilingual\\
Llama2 \cite{touvron2023llama} & 32 & 32 & 7B & 32K & 4,096 & Mixture of datasets & Primarily English\\
Llama3 \cite{touvron2023llama} & 32 & 32 & 8B & 128K & 8,192 & Mixture of datasets & Primarily English\\
\end{tblr}
\end{adjustbox}
\caption{Detailed specifications of our baseline models. Abbreviations used are: \#Layers (Number of Hidden Layers), \#Heads (Number of Attention Heads), \#Params (Total Number of Parameters), \#Vocab (Vocabulary Size), and \#MSL (Maximum Sequence Length).}
\label{tab:models_compare}
\end{table*}

\textbf{LLaMA} \cite{touvron2023llama}, or Large Language Model Meta AI, represents a significant leap in the development of foundational models for natural language inference (NLI) tasks. Introduced by Meta AI, LLaMA is designed to create a more accessible and efficient framework for researchers and developers. Available in various sizes including 7B, 13B, 33B, 65B, and 70B parameters, LLaMA caters to different computational needs and research objectives. It is trained on a diverse dataset comprising 1.4 trillion tokens from 20 languages, enabling it to perform a wide range of NLP tasks with high accuracy and efficiency. Innovations in its architecture, such as the SwiGLU activation function and rotary positional embeddings, contribute to its superior performance on NLP benchmarks.

\textbf{Mistral} \cite{jiang2023mistral}, developed by Mistral AI, stands out for its innovative approach to structured content generation and instruction-based modeling. Designed to generate high-quality, structured content similar to the functionalities offered by OpenAI models, Mistral achieves enhanced efficiency and lower resource requirements. The Mistral model utilizes mechanisms like Grouped-query Attention (GQA) and Sliding Window Attention (SWA) to achieve faster inference times and handle longer text sequences. Its ability to parse and extract information using a JSON Schema makes it particularly suited for tasks requiring structured output.

\textbf{Gemma} \cite{team2024gemma}, developed by Google DeepMind, leverages technology from the Gemini model to offer state-of-the-art, open models. It includes a 7B parameter model for GPU/TPU use and a 2B parameter model for CPU and on-device applications, both trained on up to 6 billion tokens. These models excel in language understanding, reasoning, and safety benchmarks, outperforming similarly sized open models in 11 out of 18 tasks. Key enhancements in the Gemma models include multi-query attention, RoPE embeddings, GeGLU activations, and RMSNorm for stable training. The models are rigorously evaluated through automated and human benchmarks to ensure robustness and reliability, with a strong emphasis on responsible AI practices.

The LLMs were fine-tuned using the LoRA through the Unsloth library. Detailed configuration specifics and prompting procedures are described in Section 4.2 and Appendix \ref{Prompts}, respectively.
 
\subsection{In-Context Learning Models}
In addition to fine-tuning large language models (LLMs), we rigorously assessed the in-context learning capabilities of these models through zero-shot evaluations, where models are tasked with generating accurate responses without prior specific training on examples.

Beyond the models detailed in Appendix \ref{appendix:llms}, we employed Gemini 1.5 Flash \cite{reid2024gemini}, a recent addition to the LLMs developed by Google AI. Introduced in May 2024, Gemini 1.5 Flash, part of the broader Gemini family, excels in handling multimodal tasks. Notable for its high-speed, large-scale information processing capabilities, Gemini 1.5 Flash is particularly suitable for real-time applications and environments requiring frequent updates. Despite its focus on efficiency, this model maintains robust reasoning capabilities across multiple modalities, including text, image, and audio, and supports an extensive context window of up to one million tokens. This feature is crucial for tasks that require a deep comprehension of prior information.

Furthermore, we conducted experiments on the ViFactCheck dataset using the prompt described in Appendix \ref{Prompts}, following a zero-shot approach. These experiments aimed to evaluate the ability of models to integrate and reason with various types of information without preliminary fine-tuning, showcasing its potential in real-world applications where training data may be sparse or unavailable.

\section{Prompts for Vietnames Fact-Checking} \label{Prompts}

\definecolor{mycustomcolor}{RGB}{46, 95, 127}

In this section, we outline the templates for the prompting methods used for fine-tuning and zero-shot evaluations with LLMs in the fact-checking task. The prompt structure is designed to test the ability of models to assess the veracity of a claim based on the given context or evidence. 

\begin{tcolorbox}[colback=white, colframe=mycustomcolor,boxrule=0.3mm, boxsep=1mm, left=1mm, right=1mm, top=1mm, bottom=1mm, title=Fine-Tune Instruction Prompting]
\small
You will be presented with a long context, followed by a claim. Your task is to fact-check the claim based on the provided context. You must categorize the claim into one of three categories: 

- \textbf{Support:} Choose this if the claim is true and fully supported by the context. 

- \textbf{Refute:} Choose this if the claim is false and contradicted by the context. 

- \textbf{Not Enough Information:} Choose this if the claim contains content that is not covered by the context, making it impossible to determine its accuracy.\\
\texttt{\#\#\# Context:}\\
\texttt{\#\#\# Claim:}\\
\texttt{\#\#\# Response:}
\end{tcolorbox}

\begin{tcolorbox}[colback=white, colframe=mycustomcolor,boxrule=0.3mm, boxsep=1mm, left=1mm, right=1mm, top=1mm, bottom=1mm, title=Zeroshot Prompting]
\small
Return only the label in the format: Label: Support(0), Label: Refute(1), or Label: Not Enough Information(2).

\textbf{Instructions:}

1. Fact check the claim based on the provided evidence.

2. Use the following labels:

        - \textbf{Support:} The claim is true and supported by the evidence.
        
        - \textbf{Refute:} The claim is false and contradicted by the evidence.
        
        - \textbf{Not Enough Information:} The claim contains content that is not covered by the evidence, making it impossible to determine its accuracy.
        
\textbf{Example:}\\
\texttt{Label: Support}
\end{tcolorbox}

\section{Number of Parameters}\label{sec:params}
To establish the main baseline models, we utilized several state-of-the-art methods, including a pre-trained and large language model, to support the Vietnamese Fact-Checking task. The details of each model are shown in Table \ref{tab:models_compare}.

\section{Definition and Examples of Error Analysis}
\label{sec:definition_error}
We introduce the error definition as follows and illustrate some error cases for Vietnamese fact-checking in Figure 5:

\begin{itemize}
    \item \textbf{Evidence Retrieval Failure:} Failures due to the inability of model to accurately and fully extract essential evidence from data sources (as shown in Figure \ref{fig:Retrieval_Error}).

    \begin{figure}[ht]
    \centering
    \includegraphics[width=\linewidth]{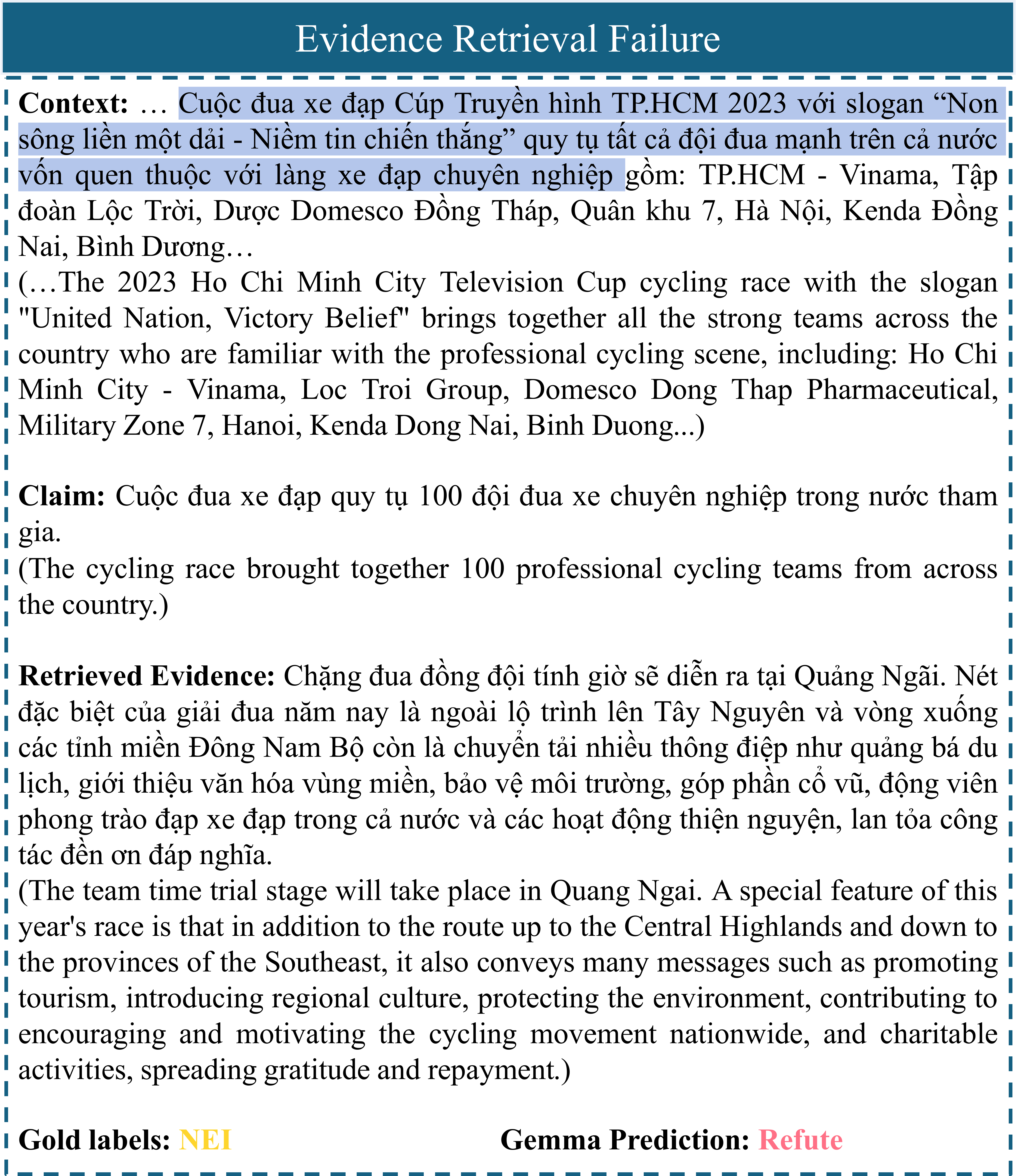}
    \caption{Examples of Evidence Retrieval Failure.}
    \label{fig:Retrieval_Error} 
    \end{figure}
    
    \item \textbf{Semantic Ambiguity:} Issues arising from context with ambiguous, verbose, or complex data, leading to interpretive difficulties (as shown in Figure \ref{fig:semantic_ambiguity}).

    \begin{figure}[ht]
    \centering
    \includegraphics[width=\linewidth]{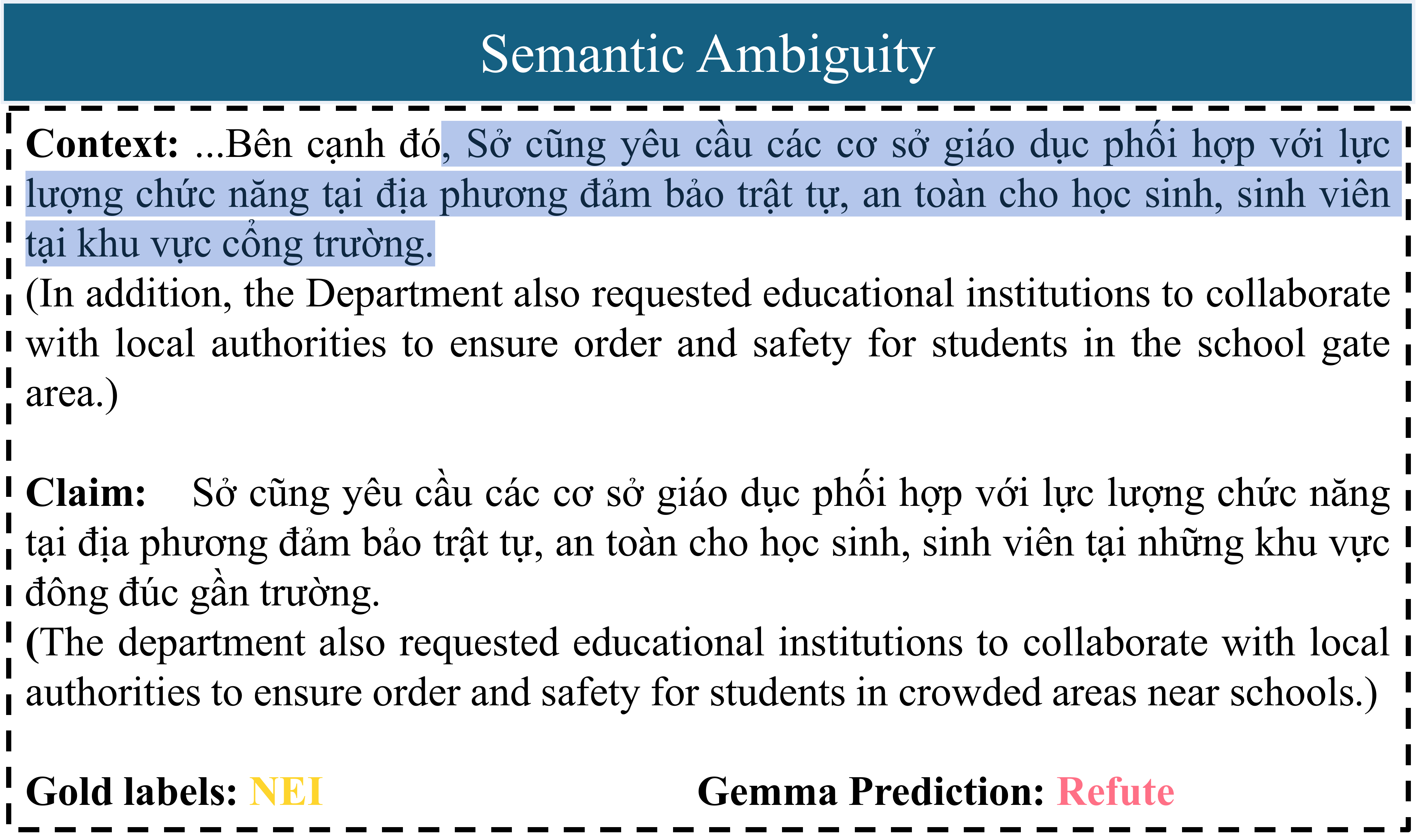}
    \caption{Examples of Semantic Ambiguity.}
    \label{fig:semantic_ambiguity} 
    \end{figure}

    \item \textbf{Inference Hallucination:} Incorrect classifications produced by the model, despite the correct extraction and availability of relevant evidence (as shown in Figure \ref{fig:hallucination}).

    \begin{figure}[ht]
    \centering
    \includegraphics[width=\linewidth]{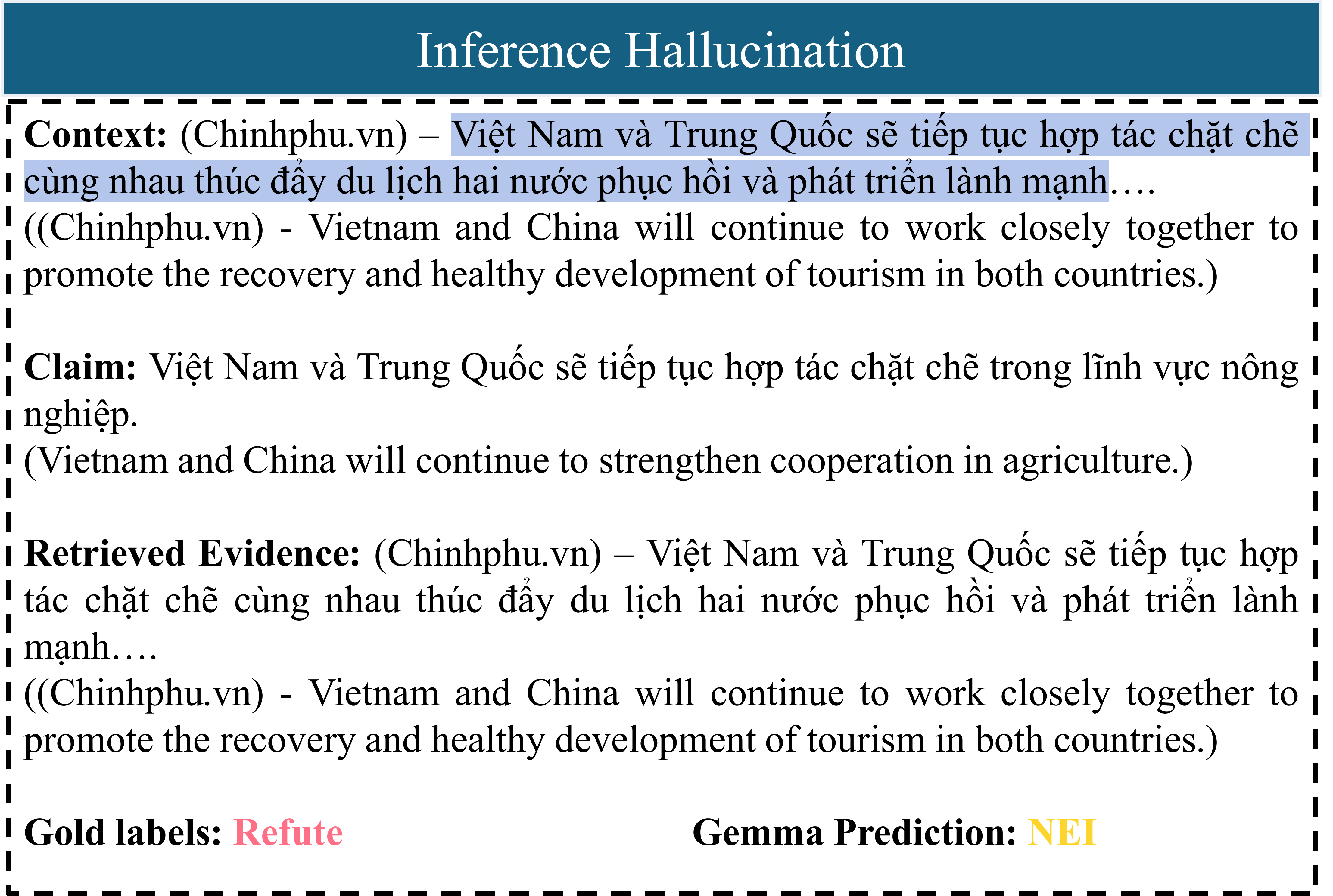}
    \caption{Examples of Inference Hallucination.}
    \label{fig:hallucination} 
    \end{figure}

    \item \textbf{Complex Inferential Chain:} Errors resulting from the necessity to synthesize insights across multiple sources or evidences through sequential reasoning (as shown in Figure \ref{fig:complex_chain}).

    \begin{figure}[ht]
    \centering
    \includegraphics[width=\linewidth]{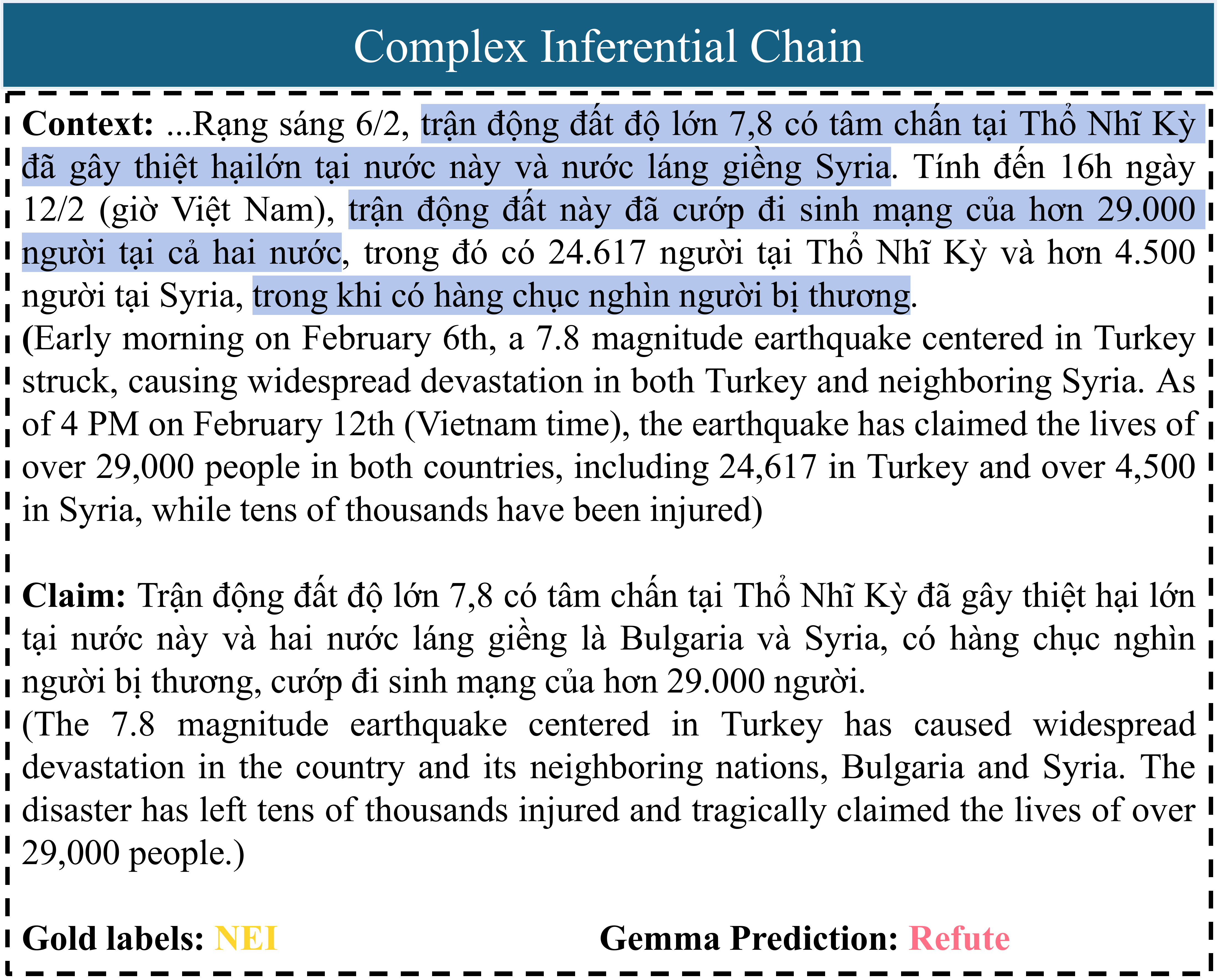}
    \caption{Examples of Complex Inferential Chain.}
    \label{fig:complex_chain} 
    \end{figure}
    
    \item \textbf{Labeling Error:} Issues stemming from inaccuracies or inconsistencies introduced during the manual data labeling process.
\end{itemize}

\section{Additional Qualitative Analysis}
To obtain insights into the performance of language models, we conducted an in-depth analysis considering various factors such as the length of the context, the topic of discussion, the volume of training data, and the duration of model training.

\begin{figure}[ht]
    \centering
    \includegraphics[width=\linewidth]{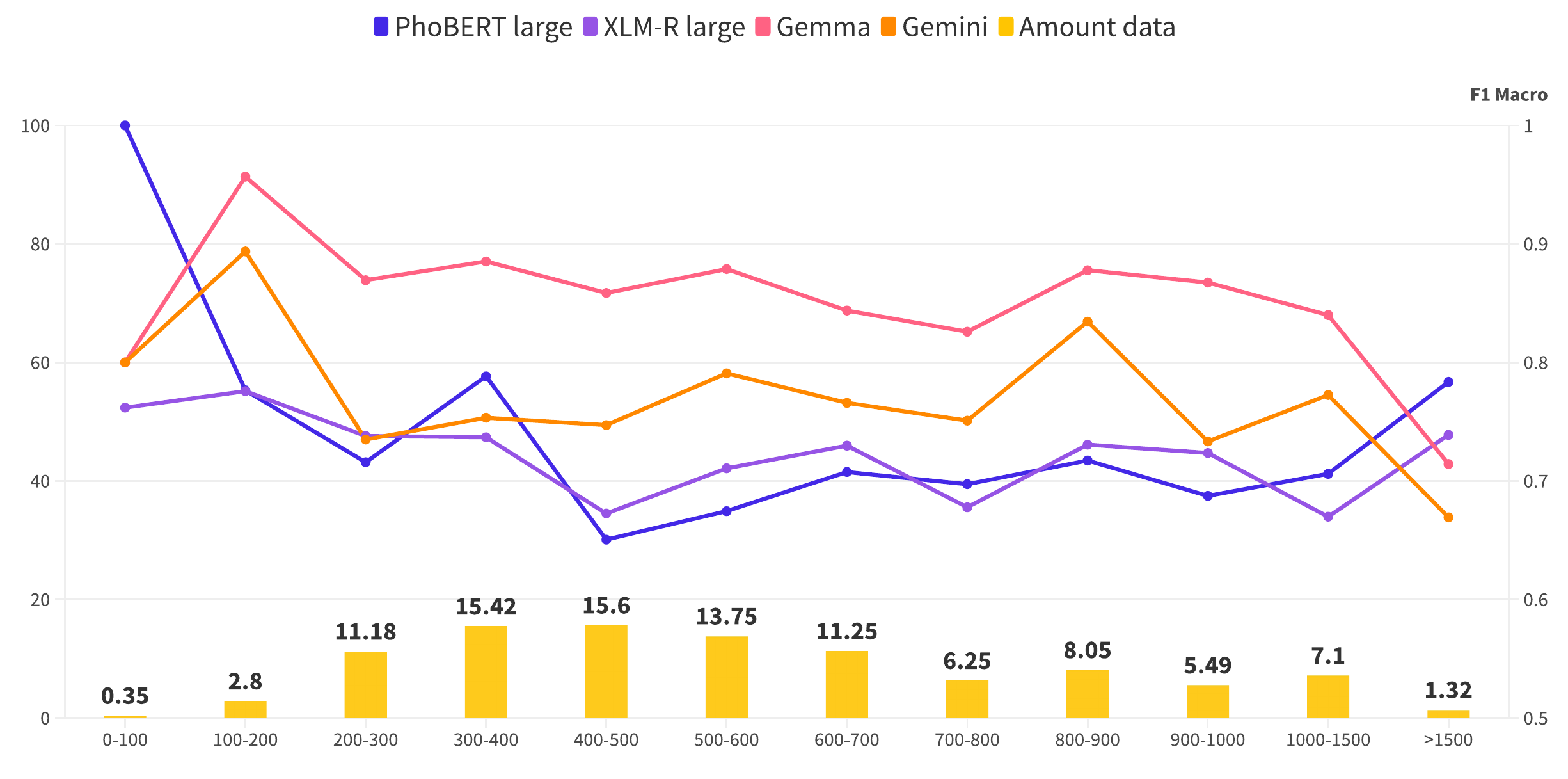}
    \caption{The effect of the length context on test set.}
    \label{fig:result-length} 
\end{figure}

\paragraph{Effects of Context Length}
We initiated our investigation by analyzing the test results with respect to the length of the context (see Figure \ref{fig:result-length}). Notably, PhoBERT$_{large}$ and XLM-R$_{large}$ perform well when analyzing shorter texts (0-100 words). However, their performance declines as text length increases, particularly in the 400-500 and 500-600 word ranges, suggesting that longer texts may pose challenges for these models. In contrast, Gemma and Gemini exhibit more consistent performance across different text lengths, showing only minor fluctuations. This stability suggests their potential suitability for tasks involving a wide range of text lengths, where maintaining accuracy is crucial. The consistent performance of Gemma and Gemini across various text lengths is particularly advantageous for fact-checking, which often involves analyzing claims of different lengths, from short social media posts to longer articles.

\begin{figure}[ht]
    \centering
    \includegraphics[width=\linewidth]{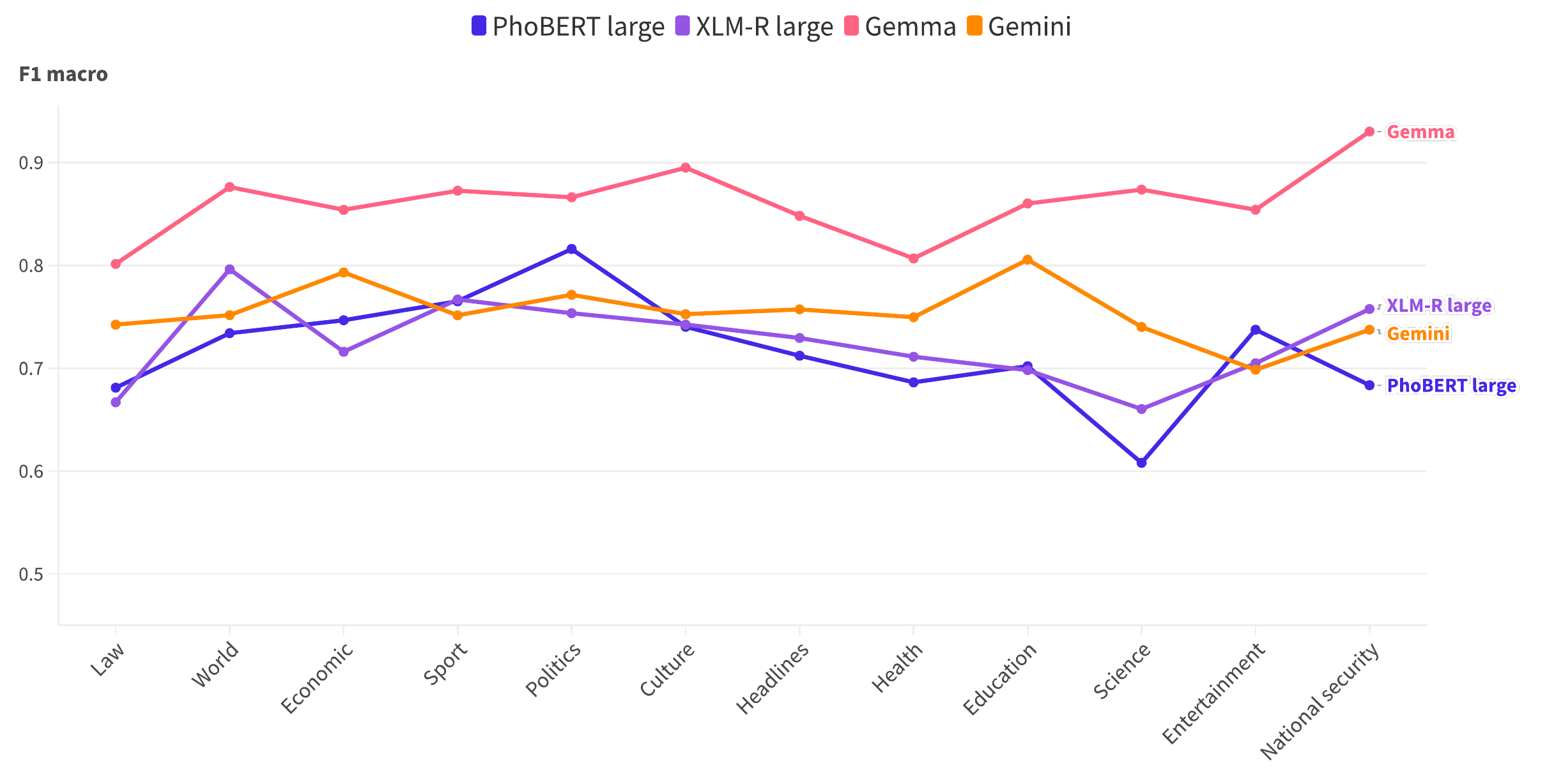}
    \caption{The effect of the topic on the test set.}
    \label{fig:topic_statistic_chart} 
\end{figure}

\paragraph{Effects of Topic}
Further analysis focused on the impact of topics on model performance, as illustrated in Figure \ref{fig:topic_statistic_chart}. Gemma consistently outperforms other models across most topics, excelling particularly in the ``Science'', ``National Security'', and "Culture" categories. Gemini generally performs the second best, closely following Gemma in most areas but showing a slight dip in ``National security'' and ``Entertainment''. While not as strong overall, PhoBERT$_{large}$ and XLM-R$_{large}$ have their strengths: PhoBERT$_{large}$ performs notably well in ``Politics'', benefiting from being pre-trained on a large Vietnamese dataset that provides an advantage in this domain due to the specific vocabulary required. Conversely, XLM-R$_{large}$ shows a relative peak in the ``World'' category, leveraging its multilingual training data to gain an advantage over monolingual models like PhoBERT$_{large}$.

Interestingly, except for Gemma, the remaining models seem to struggle with the ``Science'', ``Law'', and ``Health'' categories, indicating a potential area for improvement in Vietnamese fact-checking models. These categories require high accuracy and specialized vocabulary, which may explain the suboptimal performance of the other models. Additionally, there is a noticeable performance gap between Gemma and the other models in several topics, suggesting that the architecture or training data of Gemma might be better suited for fact-checking Vietnamese across diverse topics.



\begin{figure}[ht]
    \centering
    \includegraphics[width=\linewidth]{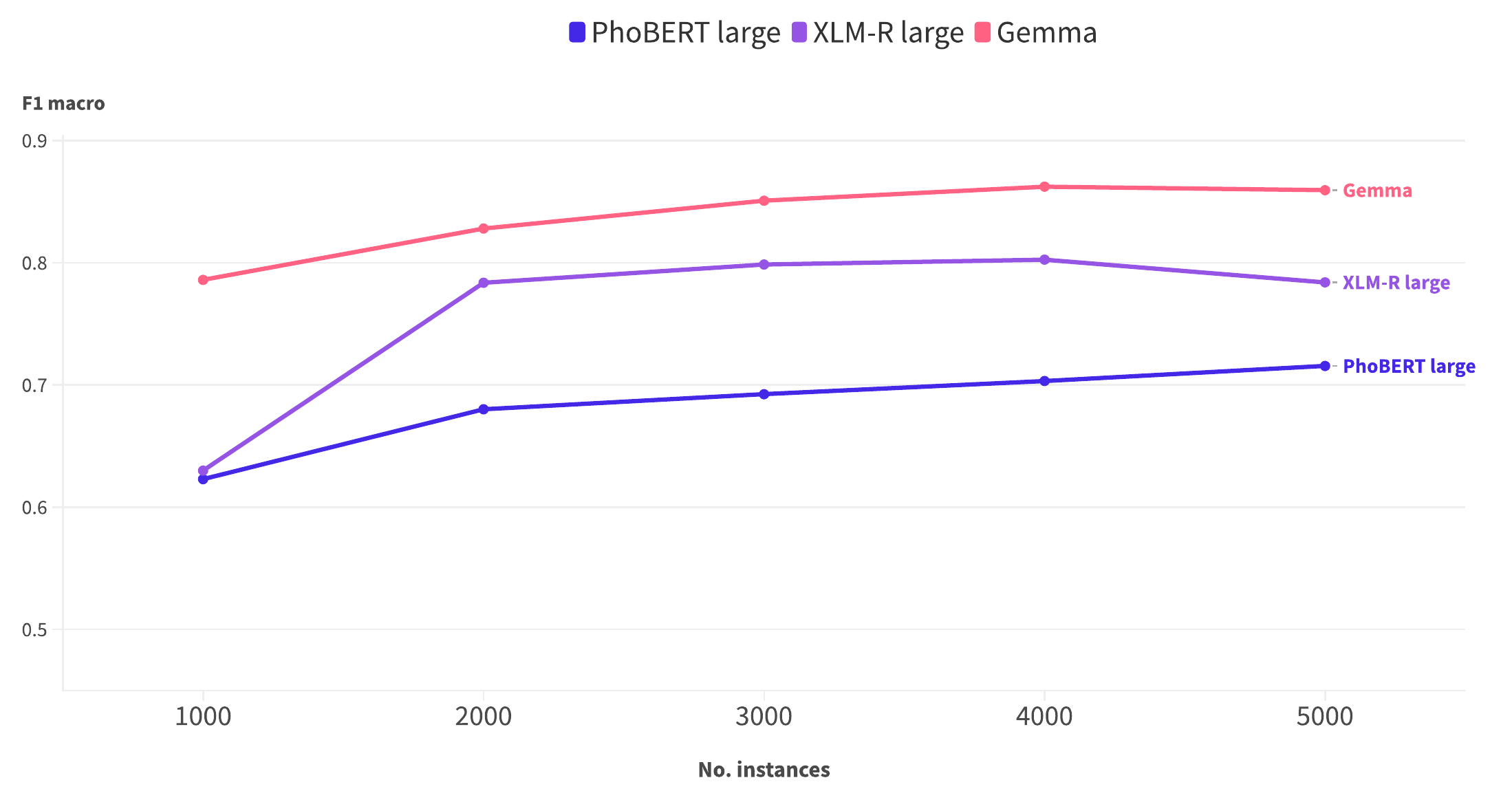}
    \caption{The impact of training data size on test set.}
    \label{fig:result_size} 
\end{figure}

\paragraph{Effects of Training Data Size} To investigate the effect of training data size on model performance, we conducted experiments with various data subsets, including those containing 1000, 2000, 3000, 4000, and 5062 data points. Figure \ref{fig:result_size} visually represents the evaluation performance across these subsets. Note that all models demonstrated improved performance as the dataset size increased. Given that Gemini is an API-based model and cannot be trained on custom datasets, it was excluded from this analysis.

Our comprehensive analysis highlights the multifaceted factors influencing model performance. Gemma consistently outperforms both PhoBERT$_{large}$ and XLM-R$_{large}$ across all training sizes. While all models exhibit improved performance with increased data, F1-score of Gemma starts higher and increases at a steeper rate, especially up to 2,000 instances. Beyond this point, the rate of improvement for all models slows, indicating diminishing returns from additional training data. This consistent superiority demonstrates effectiveness of Gemma regardless of the available training data amount.

\begin{figure*}[ht]
    \centering
         \includegraphics[width=.85\linewidth]{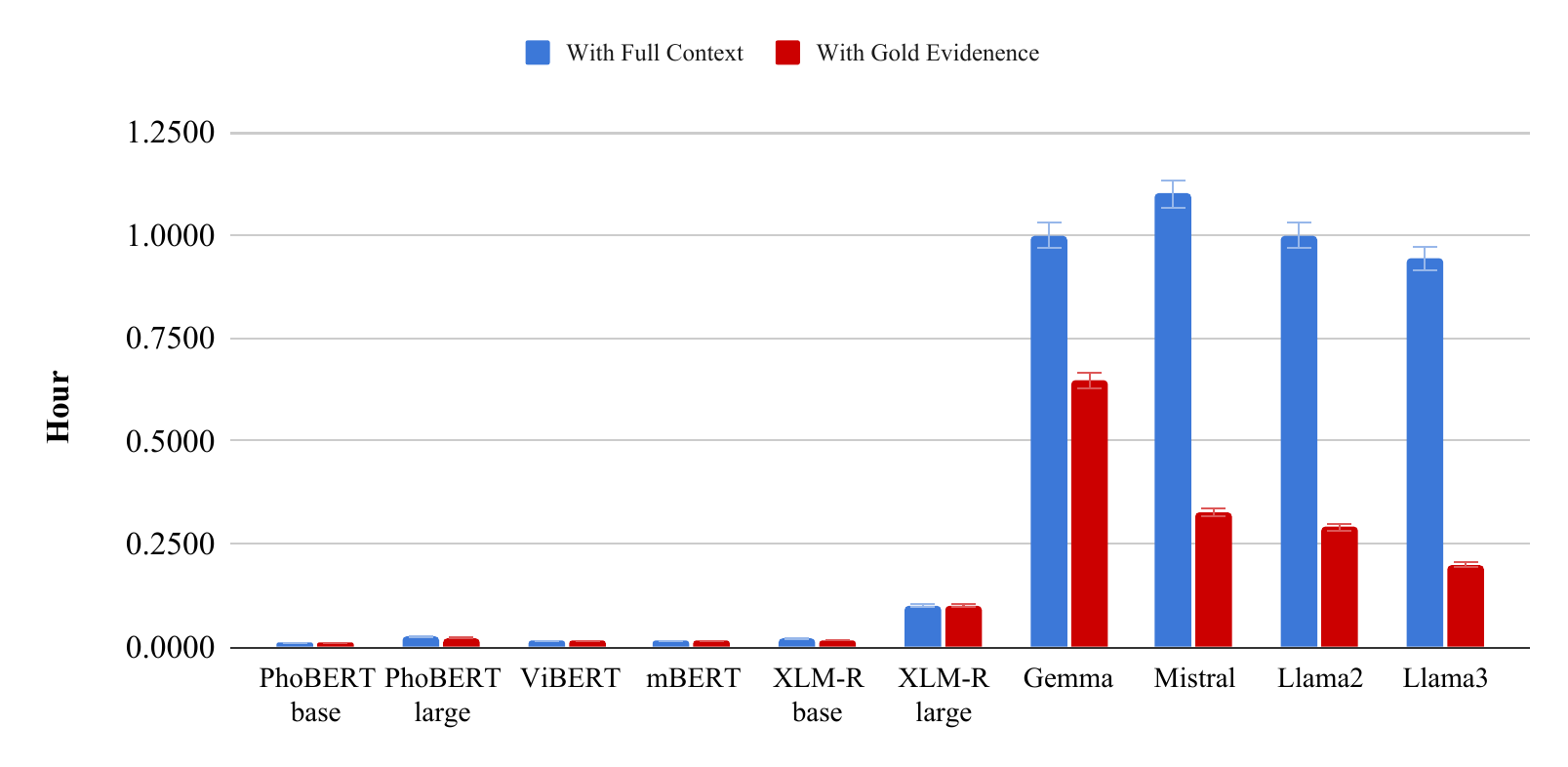}
    \caption{The comparison of training times per epoch for various baseline models.}
    \label{fig:time} 
\end{figure*}

Moreover, our findings show that increasing the size of the training data improves the performance of Vietnamese models such as PhoBERT$_{large}$, highlighting the need for a robust and diverse training dataset to achieve optimal fact-checking results.

\paragraph{Analysis of Training Time Efficiency}
Finally, Figure \ref{fig:time} illustrates the training times of various models per epoch, measured in hours. The Mistral model has the longest training time at 1.1 hours for processing Full Context (FC), indicative of its complexity and computational demands. Gemma and Llama2 each require approximately 1.0 hour, while Llama3 requires significant time as well, at 0.94 hours. These durations illustrate the intricate computations these models undertake for handling detailed and extensive contexts.

In contrast, the XLM-R$_{large}$ model, though still demanding, is more time-efficient at only 0.1 hours, likely due to its optimized large-scale architecture. The PhoBERT$_{large}$ and XLM-R$_{base}$ models show moderate training times, striking a balance between computational efficiency and performance capabilities.

Models such as mBERT, ViBERT, and PhoBERT$_{base}$ demonstrate shorter training times, ranging from 0.0083 to 0.0139 hours. These reduced durations suggest higher operational efficiency but may also indicate a lower capacity for managing complex tasks requiring extensive contextual data.

When trained with Gold Evidence, which comprises shorter and more directly relevant sentences, Gemma still requires the most time at 0.6467 hours, although this is significantly less than with Full Context. Mistral, Llama2, and Llama3 also exhibit reduced training times at 0.32, 0.29, and 0.20 hours, respectively. This indicates that models can achieve greater efficiency when provided with concise and pertinent training data.

This analysis underscores the trade-offs between training time, model complexity, and performance, highlighting the substantial computational demands placed on advanced models to achieve high performance in Vietnamese fact-checking tasks. The reduced training times with Gold Evidence further emphasize the potential efficiency gains from using relevant training inputs.

In conclusion, our analysis elucidates the multifaceted effects of dataset characteristics and training time on the performance of language models. Larger and more diverse datasets generally improve model accuracy, particularly in specialized applications like fact-checking. However, the efficiency of model training also plays a critical role, as faster training can lead to quicker deployment and adaptation in dynamic environments. The results underscore the importance of optimizing both the data input and model architecture to achieve the best balance between performance and efficiency, which is crucial to develop robust AI systems capable of handling the intricacies of language-based tasks.

\section{Scientific Artifacts}
The licenses for all the models and software used in this paper are listed in parentheses: Beautiful Soup 4 (MIT License), Selenium (Apache License 2.0), Fleiss Kappa (BSD License), mBERT (Apache License 2.0), ViBERT (Apache License 2.0), PhoBERT (MIT License), XLM-R (Apache License 2.0), VnCoreNLP (Apache License 2.0), Unsloth (Apache License 2.0), LoRa (Apache License 2.0), F1-score (BSD License), BM25 ( MIT License), SBERT (Apache License 2.0), Gemma (Apache License 2.0), Mistral (Apache License 2.0), Llama3 (Apache License 2.0), Llama2 (Apache License 2.0) Gemini 1.5 Flash (Proprietary License), Label Studio (Apache License 2.0)


\end{document}